\documentclass[sigconf]{acmart}

\usepackage{newtxmath}   

\AtBeginDocument{%
  }

\setcopyright{acmlicensed}
\copyrightyear{2025}
\acmYear{2025}
\setcopyright{acmlicensed}\acmConference[KDD '25]{Proceedings of the 31st ACM SIGKDD Conference on Knowledge Discovery and Data Mining V.2}{August 3--7, 2025}{Toronto, ON, Canada}
\acmBooktitle{Proceedings of the 31st ACM SIGKDD Conference on Knowledge Discovery and Data Mining V.2 (KDD '25), August 3--7, 2025, Toronto, ON, Canada}
\acmDOI{10.1145/3711896.3737018}
\acmISBN{979-8-4007-1454-2/2025/08}

\usepackage{multirow}
\usepackage{graphicx}
\usepackage{booktabs}
\usepackage{tabularx}
\usepackage{amssymb}
\usepackage{pifont}
\usepackage{makecell}

\begin{document}

\title{Learnable-Differentiable Finite Volume Solver for Accelerated Simulation of Flows}

\author{Mengtao Yan}
\email{mengtaoyan@ruc.edu.cn}
\orcid{0000-0001-7583-0277}
\affiliation{%
  \institution{Renmin University of China}
  \city{Beijing}
  \country{China}
}

\author{Qi Wang}
\email{qi_wang@ruc.edu.cn}
\orcid{0009-0009-4712-3474}
\affiliation{%
  \institution{Renmin University of China}
  \city{Beijing}
  \country{China}
}

\author{Haining Wang}
\email{wanghaining5@hisilicon.com}
\orcid{0000-0002-7412-3797}
\affiliation{%
  \institution{Huawei Technologies Ltd.}
  \city{Hangzhou}
  \country{China}
}

\author{Ruizhi Chengze}
\email{chengzeruizhi@huawei.com}
\orcid{0009-0002-5736-0165}
\affiliation{%
  \institution{Huawei Technologies Ltd.}
  \city{Shanghai}
  \country{China}
}

\author{Yi Zhang}
\email{zhangyi430@huawei.com}
\orcid{0000-0003-3487-7073}
\affiliation{%
  \institution{Huawei Technologies Ltd.}
  \city{Hangzhou}
  \country{China}
}

\author{Hongsheng Liu}
\email{liuhongsheng4@huawei.com}
\orcid{0000-0003-0509-7967}
\affiliation{%
  \institution{Huawei Technologies Ltd.}
  \city{Shanghai}
  \country{China}
}

\author{Zidong Wang}
\email{wang1@huawei.com}
\orcid{0009-0007-4524-1384}
\affiliation{%
  \institution{Huawei Technologies Ltd.}
  \city{Hangzhou}
  \country{China}
}

\author{Fan Yu}
\email{fan.yu@huawei.com}
\orcid{0009-0001-2189-351X}
\affiliation{%
  \institution{Huawei Technologies Ltd.}
  \city{Hangzhou}
  \country{China}
}

\author{Qi Qi}
\email{qi.qi@ruc.edu.cn}
\orcid{0000-0001-9192-8928}
\affiliation{%
  \institution{Renmin University of China}
  \city{Beijing}
  \country{China}
}

\author{Hao Sun}
\email{haosun@ruc.edu.cn}
\authornote{Corresponding author.}
\orcid{0000-0002-5145-3259}
\affiliation{%
  \institution{Renmin University of China}
  \city{Beijing}
  \country{China}
}

\renewcommand{\shortauthors}{Mengtao Yan et al.}
\newcommand{\qwedit}[1]{{\color{red} #1}}
\newcommand{\qwcomment}[1]{{\color{green} [qw: #1]}}
\begin{abstract}
  Simulation of fluid flows is crucial for modeling physical phenomena like meteorology, aerodynamics, and biomedicine. Classical numerical solvers often require fine spatiotemporal grids to satisfy stability, consistency, and convergence conditions, leading to substantial computational costs. Although machine learning has demonstrated better efficiency, they typically suffer from issues of interpretability, generalizability, and data dependency. Hence, we propose a learnable and differentiable finite volume solver, called LDSolver, designed for efficient and accurate simulation of fluid flows on spatiotemporal coarse grids. LDSolver comprises two key components: (1) a differentiable finite volume solver, and (2) an learnable module providing equivalent approximation for fluxes (derivatives and interpolations), and temporal error correction on coarse grids. Even with limited training data (e.g., only a few trajectories), our model could accelerate the simulation while maintaining a high accuracy with superior generalizability. Experiments on different flow systems (e.g., Burgers, decaying, forced and shear flows) show that LDSolver achieves state-of-the-art performance, surpassing baseline models with notable margins.
\end{abstract}




\maketitle

\section{Introduction}
\label{submission}

Fluid flow simulation is crucial for various applications, including weather and climate forecasting~\cite{bauer2015quiet,schneider2017climate}, aerospace engineering~\cite{slotnick2014cfd}, automotive design~\cite{kobayashi2009cfd}, and biomedicine~\cite{basri2016computational}. These flows are generally governed by nonlinear partial differential equations (PDEs), particularly the Navier-Stokes equations (NSE). However, the accuracy of traditional solvers heavily depends on spatiotemporal resolution~\cite{AndersonWendt1995,MoukalledManganiDarwish2016,ZienkiewiczTaylorZhu2005,KarniadakisSherwin2005}. While fine grids are essential for ensuring convergence and solution validity, they result in extremely high computational costs. For instance, accurately simulating the flow field around an aircraft wing often requires tens to hundreds of millions of grid cells~\cite{shang2009computational}. Such simulations typically take several days to a week to complete on high-performance computing clusters, utilizing hundreds of CPUs or multiple GPUs for parallel processing. Furthermore, any changes in initial or boundary conditions (IC/BCs) or design parameters require a full re-simulation. Additionally, a significant limitation of existing solvers, such as OpenFOAM~\cite{jasak2007openfoam} and Fluent~\cite{matsson2023introduction}, is their inherent non-differentiability, which makes integration with neural networks challenging.

Recent advances in machine learning (ML) have demonstrated significant potential for accelerating the simulation of complex flows~\cite{brunton2020machine}. Existing ML-driven approaches can be broadly categorized into three groups. First, data-driven methods train models using supervised learning on pre-existing datasets. By bypassing the time-resolution limitations of traditional solvers, these methods often ignore underlying partial differential equation (PDE) constraints. As a result, they frequently suffer from poor interpretability, limited generalizability, and a strong reliance on large, labeled datasets~\cite{LuJinPangZhangKarniadakis2021,LiKovachkiAzizzadenesheliLiuBhattacharyaStuartAnandkumar2021}. To address these issues, integrating physical information into the training process has been explored. Second, physics-informed neural networks (PINNs) incorporate PDEs and IC/BCs as regularization terms in the loss function, imposing soft constraints. However, PINNs face challenges related to training convergence, solution accuracy, and robust generalization~\cite{RaissiPerdikarisKarniadakis2019, RaissiYazdaniKarniadakis2020,WangKashinathMustafaAlbertYu2020,ZhangLiuSun2020,ChenLiuSun2021,RaoSunLiu2021,TangZhaiWanYang2024}. Third, another category explicitly embeds physical information directly into the network architecture as hard constraints. For example, physics-encoded recurrent convolutional neural networks (PeRCNN) use neural networks within a finite difference (FD) framework to refine derivative approximations, improving generalization without extensive hyperparameter tuning. However, PeRCNN can still exhibit instability in long-term predictions~\cite{RaoSunLiu2021}. Learning interpolation (LI) models~\cite{Kochkov2021} and TSM~\cite{sun2023neural}, operating within a finite volume framework, employ neural networks to learn the most resolution-sensitive components of traditional solvers. This enables faster simulations on coarser grids while maintaining accuracy. Nevertheless, due to the non-trainable nature of other numerical components, these models still require substantial labeled data for accurate predictions.

\begin{figure}[t!]
\vskip 0in
\begin{center}
\centerline{\includegraphics[width=0.48\textwidth]{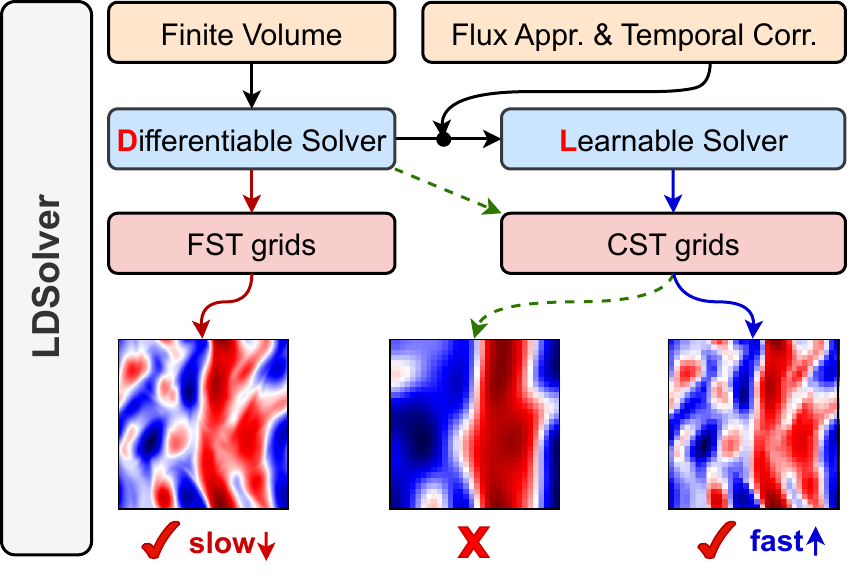}}
\caption{Overview of LDSolver. It comprises a differentiable traditional solver integrated with finite volume methods and a learnable solver incorporating flux approximation (Appr.) and temporal correction (Corr.) blocks. \textcolor{red}{Red} arrows indicate simulations by the traditional solver on fine spatiotemporal (FST) grids, while \textcolor{blue}{blue} arrows represent accelerated simulations by the learnable solver on coarse spatiotemporal (CST) grids. \textcolor{green}{Green} arrows highlight erroneous solutions produced by the traditional solver on CST grids.}
\Description{Diagram of LDSolver architecture with red, blue, and green arrows indicating different solver behaviors on fine and coarse spatiotemporal grids.}
\label{fig:f1_Solver}
\end{center}
\vskip 0in
\end{figure}

To address these challenges, we introduce LDSolver, a method designed for stable and high-fidelity long-term predictions, even with limited training data. Our approach begins by developing a differentiable numerical solver capable of generating high-fidelity solutions on fine grids. Training datasets are then created by downsampling these high-resolution solutions to the desired coarse grid resolutions. To correct accumulated errors on coarse grids, we employ neural networks to optimize derivatives and interpolations estimation mechanisms while strictly preserving the equation form.  A temporal correction block further refines predictions over time, inspired by recent works~\cite{poli2020hypersolvers,berto2022neural}. Motivated by advances in differentiable physics~\cite{um2020solver,holl2024bf}, LDSolver is constructed on a fully differentiable solver backbone. As shown in Figure~\ref{fig:f1_Solver}, our main contributions are summarized as follows:

\begin{itemize}
\item We propose a novel learnable and differentiable finite volume solver (LDSolver) for accelerated flow simulations. This hybrid approach integrates traditional numerical solvers with machine learning, enabling high-fidelity simulations on fine grids and efficient learning and prediction on coarse grids, achieving strong generalization with limited data.
\item We introduce a flux block and a temporal correction block to estimate derivatives and interpolations on coarse grids, effectively reducing accumulated errors during long-time iterations on coarse grids.
\item LDSolver demonstrates significant improvements across various complex flows, including Burgers, decaying, forced, and shear flows. Compared to baseline models, it achieves at least a 50\% reduction in mean absolute error (MAE) while maintaining high computational efficiency.
\end{itemize}

\section{Related Work}

\textbf{Numerical Solvers:} 
\label{sec:numerical_solver}
Traditionally, fluid dynamic systems are typically solved using numerical techniques, such as finite difference methods (FDM)~\cite{AndersonWendt1995}, finite volume methods (FVM)~\cite{MoukalledManganiDarwish2016}, finite element methods (FEM)~\cite{ZienkiewiczTaylorZhu2005}, and spectral methods~\cite{KarniadakisSherwin2005}. Among these, FVM is a widely adopted numerical approach in computational fluid dynamics (CFD). By discretizing the integral form of the governing equations within control volumes, it ensures local conservation, robust stability, and the capability to handle complex geometries and boundary conditions. These methods generally require fine spatio-temporal discretizations to meet stability, consistency, and convergence criteria. However, the computational cost remains prohibitively high for large-scale simulations or inverse analyses.
 
\textbf{Pure Data Driven Methods:} 
Recently, deep learning has been increasingly applied to solving partial differential equations (PDEs). Common approaches include methods based on convolutional neural networks (CNNs)~\cite{Bar-Sinai2019,yan2022predicting}, U-Nets~\cite{gupta2023towards}, ResNets~\cite{Lu2018}, graph neural networks (GNNs)~\cite{Sanchez-Gonzalez2020,Pfaff2021}, and Transformer-based models~\cite{Janny2023, Li2024, Wu2024}. Additionally, neural operator methods, such as DeepONet~\cite{LuJinPangZhangKarniadakis2021}, multiwavelet-based models (MWTs)~\cite{Gupta2021}, Fourier Neural Operators (FNO)~\cite{2021Fourier}, and their variants~\cite{Tran2022,Li2024b,Goswami2022}, directly learn mappings between functions, making them particularly suitable for building surrogate models of PDE systems. These methods are typically purely data-driven, relying on extensive data, and they often suffer from limited interpretability and generalizability.

\textbf{Physics-aware Learning Methods:} 
Physics-aware learning methods have demonstrated significant potential in modeling spatiotemporal dynamics with limited data. These methods can be broadly categorized into two groups based on how they incorporate physics: (1) Physics-informed methods utilize partial differential equations (PDEs) and initial/boundary conditions (I/BCs) as soft constraints, often integrated into the loss function for regularization. Examples include PINNs~\cite{RaissiPerdikarisKarniadakis2019,RaissiYazdaniKarniadakis2020,WangKashinathMustafaAlbertYu2020,TangZhaiWanYang2024}, PhyGeoNet~\cite{Gao2021}, PhyCRNet~\cite{Ren2022}, and PhySR~\cite{Ren2023}. (2) Physics-encoded methods treat PDE structures (such as equations, I/BCs) as hard constraints, directly embedding them into the neural network architecture. Examples include PeRCNN~\cite{RaoRenLiuSun2022,Rao2023}, FluxGNN~\cite{horie2024graph}, TiGNN~\cite{Hernandez2023}, EquNN~\cite{Wang2021}, and PDE-Net models~\cite{Long2018,Long2019}, which design convolutional kernels to approximate differential operators and simulate system dynamics.

\textbf{Hybrid Physics-ML Methods:} 
Hybrid Physics-ML methods leverage neural networks to correct errors from classical numerical simulators, typically applied to low-resolution models. These methods can be integrated with various traditional numerical techniques, such as finite difference methods (e.g., PPNN~\cite{Liu2024}, P$^2$C$^2$Net~\cite{wang2024p}, learned numerical discretizations~\cite{Zhuang2021}), finite volume methods (e.g., LI~\cite{Kochkov2021}, TSM~\cite{Sun2023}), and spectral methods (e.g., machine learning-enhanced spectral solvers~\cite{Dresdner2023}). By operating on coarse grids, they achieve significant simulation speedups while maintaining reasonable accuracy. However, due to the non-trainable nature of most numerical components, these models often require substantial labeled data to ensure accurate predictions.

\begin{figure*}[t!]
  \centering
  \includegraphics[width=0.97\linewidth]{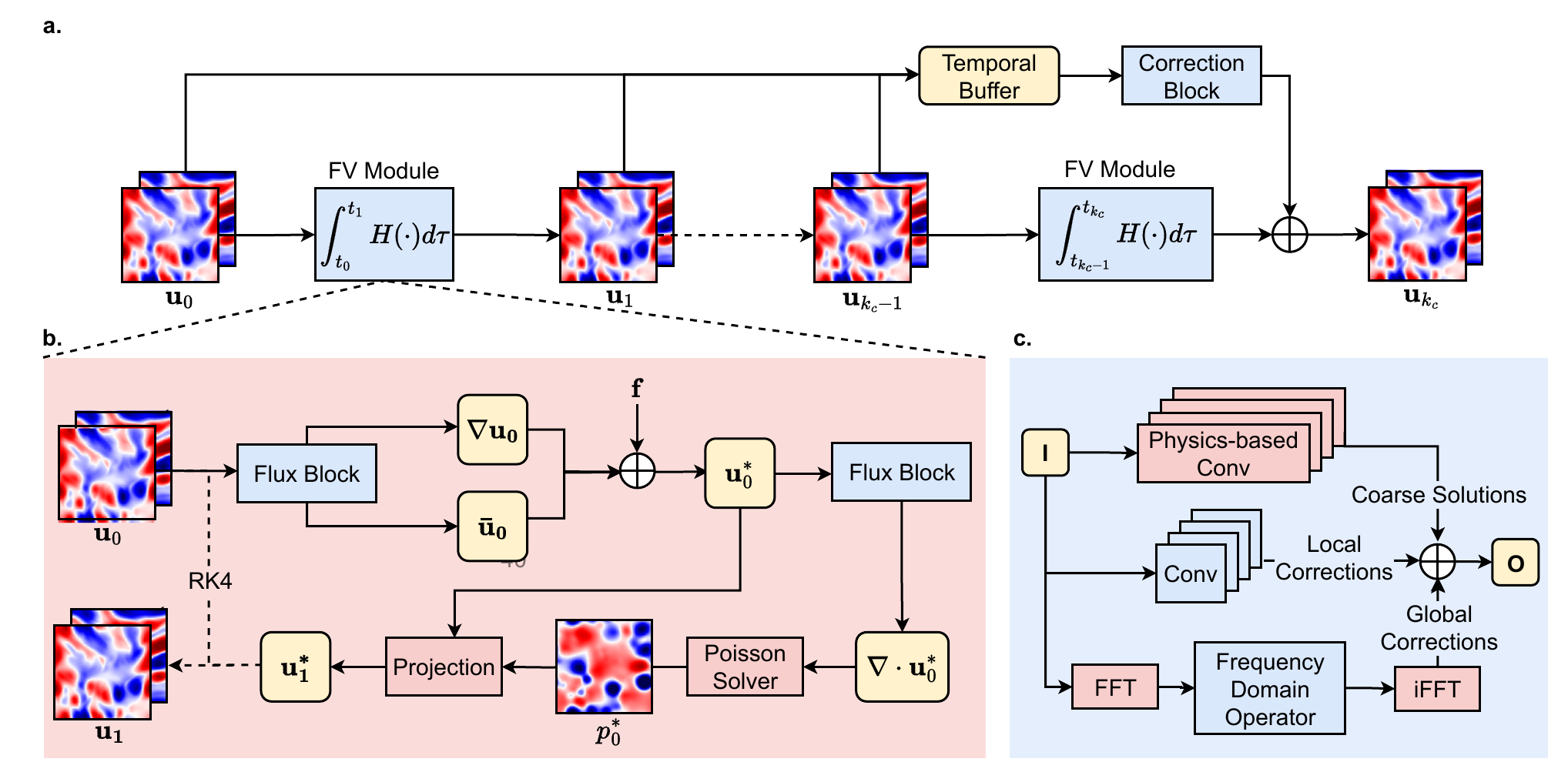}
  \vspace{-8pt}
  \caption{Architecture of the learnable differentiable solver (LDSolver), illustrated using forced flow as an example. LDSolver consists of a finite volume (FV) module with a flux block and a temporal correction block. (a) Overall architecture; (b) Finite volume module (FV module); (c) Flux block.}
  \Description{Illustration of LDSolver architecture, showing the overall structure, finite volume module, and flux block, in a forced flow scenario.}
  \label{fig:network}
\end{figure*}

\section{Methodology}
\subsection{Problem Statement}
\label{eq: conservation PDE}
Generally, the dynamics of fluid systems are governed by conservation-form partial differential equations (PDEs) of the following type:
\begin{equation}
    \frac{\partial \mathbf{u}}{\partial t} + \nabla \cdot \mathbf{F}(\mathbf{u, \mu}) = \mathbf{S}(\mathbf{u}),
    \label{eq:conservaion_pde}
\end{equation}
where $\textbf{u}$ denotes the flow field, $\mathbf{F}$ denotes the flux function, $\mathbf{S}$ represents source terms, and ${\mu}$ is the PDE parameters (e.g., Reynolds number or diffusion coefficient). This formulation provides a unified framework for the governing equations of diverse flow systems, including Burgers, decaying, forced, and shear flows. For instance, the Burgers' equation, with ($\mathbf{S} = 0, \mathbf{F} = \mathbf{u}^2/2 - \nu\nabla\mathbf{u}$), can be expressed within this framework.

Our goal is to develop a learnable solver trained on limited data to accelerate simulations of various flow systems on coarse grids. The proposed solver aims to retain the versatility of traditional solvers in simulating diverse flows while addressing their reliance on fine grids. By optimizing flux approximations and temporal corrections, the solver achieves significant computational efficiency and ensures generalizability across flow systems, including variations in initial conditions (ICs) and PDE parameters (${\mu}$).

\subsection{Main architecture of LDSover}

The architecture of LDSolver is depicted in Figure~\ref{fig:network}. Additionally, commonly used symbols are provided in Appendix Table~\ref{tab:ablation}. Taking the NSE as an example, the solver comprises two main modules: (1) an upper pathway for the temporal correction block, and (2) a lower pathway for the finite volume (FV) module. The solver is designed to be flexibly adaptable for handling various equations.

The upper path (\textit{Temporal Correction Block}) illustrates the process of updating the flow field from the initial field $\mathbf{u}_0$ to $k_c$-th field $\mathbf{u}_{k_c}$, where $k_c$ is the temporal correction interval. The updates for the first $k_c-1$ steps are performed through time integration in the FV module, with the updated field $\mathbf{u}$ stored in the temporal buffer after each step. At the $k_c$-th step, the velocities $\mathbf{u}_{[0:k_{c-1}]}$ from the previous $k_{c-1}$ steps are fed into the temporal correction block to compute the accumulated temporal error. The update of $\mathbf{u}_{k_c}$ is then obtained by combining the output of the FV module with this error compensation. This process repeats until the final time step $N$ is reached, effectively correcting the temporal error every $k_c$ steps.

The lower pathway (\textit{FV module}) updates the solution across adjacent time steps using finite volume spatial discretization and a fourth-order Runge-Kutta (RK4) time integrator. Starting from the initial field $\mathbf{u}_0$, the flux block computes $\boldsymbol{\nabla} \mathbf{u}_0$, representing the derivative values at the interfaces, and $\bar{\mathbf{u}}_0$, the average value. An intermediate velocity field $\mathbf{u}_0^*$ is then obtained by incorporating the external force term $\mathbf{f}$. Next, the divergence $\boldsymbol{\nabla} \cdot \mathbf{u}_0^*$ is computed by the flux block, and a Poisson solver determines the intermediate pressure field $p_0^*$. Finally, a projection step updates the velocity to $\mathbf{u}_1^*$, completing one RK4 sub-step. This process is repeated for the subsequent 4 time steps to obtain the next field $\mathbf{u}_1$.

In summary, given the initial field $\mathbf{u}_0$, the computation of $\mathbf{u}_k$ for $k \in [1, N]$ can be defined as follows:
\begin{equation}
\mathbf{u}_k =
\begin{cases} 
\mathcal{FV}(\mathbf{u}_{k-1}) + \mathcal{TC}(\textbf{u}_0, \dots, \textbf{u}_{k-1}), & \text{if } k = M k_c, \\
\mathcal{FV}(\mathbf{u}_{k-1}), & \text{otherwise},
\end{cases}
\label{eq:u_k}
\end{equation}
where $\mathcal{FV}(\mathbf{u}_{k-1})$ represents the update by the Finite Volume (FV) module, $\mathcal{TC}(\mathbf{u}_{k-1})$ denotes the temporal correction (TC) block, and $M$ is a positive integer. The whole solution $\mathbf{u}_{[0, N]}$ is obtained by combining the outputs of both modules at every $k_c$ step. For other time steps, the solution is updated exclusively by the FV module.

\subsection{Learnable Finite Volume Solver}

As previously mentioned in Section~\ref{sec:numerical_solver}, FVM offers inherent advantages in computational fluid dynamics~\cite{MoukalledManganiDarwish2016}. Therefore, we adopt it as the foundational framework. For a control volume $V^{ij} \to \mathbb{R}^2$ at the $i,j$-th grid coordinate, integrating Eq.~\ref{eq:conservaion_pde} and applying the divergence theorem yields (for simplicity, $i,j$ is omitted below):
\begin{equation}
\frac{\partial}{\partial t} \int_{V} \mathbf{u} \, dV = \oint_{\partial V} \mathbf{F}(\mathbf{u}) \cdot \mathbf{n} \, dA + \int_{V} \mathbf{S} \, dV,
\label{eq:integrate_form}
\end{equation}
where $\partial V$ and $\mathbf{n}: \partial V \to \mathbb{R}^2$ represent the boundary of the cell $V$ and the unit normal vector field pointing outward, respectively. Taking the governing equation of flow, i.e., the NSE (Appendix Eq.~\ref{eq:momontum_eq}), as an example, Eq.~\ref{eq:integrate_form} can be discretized using linear interpolation as:
\begin{equation}
\frac{\partial}{\partial t} (V\mathbf{u}) = -\sum_{f\in\partial V} A_f\left(\mathbf{F}_d + \mathbf{F}_a + \mathbf{F}_p\right)\cdot\mathbf{n}_f + \mathbf{S}V,
\end{equation}
where  $A_f$ denotes the surface area of the control volume $V$, $\mathbf{F}_d$ is the dissipation flux, $\mathbf{F}_a$ represents the advection flux, and $\mathbf{F}_p$ corresponds to the pressure flux.

On coarse spatial grids, numerical flux computations suffer from reduced accuracy. As detailed in Appendix Section \ref{appendix:Discretization}, accurate discretization of dissipation fluxes ($\textbf{F}_d$) requires precise estimation of interfacial derivatives, while advection fluxes ($\textbf{F}_a$) depend on accurate interpolation of interfacial velocities. Pressure flux ($\textbf{F}_p$) computation, typically involving the solution of a Poisson equation, also necessitates accurate evaluation of the velocity field divergence (a derivative operation). On coarse temporal grids, large time intervals can also lead to significant error accumulation. To mitigate these errors, we introduce a flux block and a temporal correction block within a finite volume framework.

\subsection{Flux Block}

The flux block is designed to compute derivatives and interpolations, combining three complementary components through an integrated operator (Figure~\ref{fig:network}c):
\begin{equation}
\mathcal{O}^{(m)}({\textbf{u}}) = \underbrace{\mathcal{O}_P^{(m)}}_{\text{Physics}} + \underbrace{\mathcal{O}_L^{(m)}}_{\text{Learnable}} + \underbrace{\mathcal{O}_F^{(m)}}_{\text{Frequency}},
\end{equation}
where $m \in \{d,i\}$ distinguishes derivative ($d$) and interpolation ($i$) operations. Each component maintains dimension-specific parameters without weight sharing.

\textbf{Physics-based Conv.} The physics-based convolution is designed to rigorously preserve physical consistency through non-trainable 2D convolutions rooted in numerical analysis foundations:
\begin{equation}
\mathcal{O}_P^{(m)}({\textbf{u}}) = \mathbf{W}_P^{(m)} \ast \textbf{u},\quad \mathbf{W}_P^{(m)} \in \mathbb{R}^{5\times4},
\end{equation}
where $\mathcal{O}_P^{(m)}$ denotes the physics-based operator, $\mathbf{W}_P^{(m)}$ represents the unlearnable convolution kernel and $\textbf{u}$ indicates the input flow field. The $5\times4$ kernels implement finite difference operators~\cite{Long2018} - central differencing for derivatives ($m=d$) and linear interpolation for flux reconstruction ($m=i$), ensuring strict adherence to numerical conservation laws. This physics-based design guarantees solution boundedness and discrete conservation (complete kernel configurations in Appendix Eq.~\ref{eq:physics_kernels}).

\textbf{Learnable Conv.} The learnable convolution enhances local discretization accuracy through parameterized convolutions to address coarse-resolution limitations:
\begin{equation}
\mathcal{O}_L^{(m)}(\textbf{u}) = \mathbf{W}_L^{(m)} \ast \textbf{u},\quad \mathbf{W}_L^{(m)} \in \mathbb{R}^{5\times4},
\end{equation}
where $\mathcal{O}_L^{(m)}$ represents the learnable operator, $\mathbf{W}_L^{(m)}$ denotes the trainable convolution kernel, and $\textbf{u}$ indicates the input flow field. The $5\times4$ kernels, initialized with  minor random values, are optimized during training to adjust stencil weights. This adaptive mechanism preserves structured connectivity of finite volume discretizations while compensating for truncation errors.

\textbf{Frequency Domain Operator.} The frequency domain operator captures global features of the flow field. We employ a Fourier neural operator (FNO)~\cite{2021Fourier} to implement this component:
\begin{equation}
\mathcal{O}_F^{(m)}(\textbf{u}) = Q\left(\mathcal{F}^{-1}\left(R_L \cdot \mathcal{F}(\textbf{u})\right)\right),
\end{equation}
where $\mathcal{O}_F^{(m)}$ denotes the frequency domain operator, $\mathcal{F}$ and $\mathcal{F}^{-1}$ represent the forward and inverse Fourier transforms, respectively, $R_L$ denotes the frequency-space correction layers, and $Q$ is a learnable projection mapping spectral features back to physical space. The architecture first decomposes the flow field into spectral components via $\mathcal{F}$, applies $L$-layer frequency-space corrections, and then reconstructs the enhanced features through $\mathcal{F}^{-1}$. The projection $Q$ ensures dimensional consistency, effectively addressing multiscale resolution challenges (implementation details in Appendix~\ref{appendix:fno}). Network hyperparameters are documented in Appendix Table~\ref{tab:network_hyperparameters}.

\subsection{Temporal Correction Block}

\begin{table*}[t!]
  \caption{Summary of datasets and training configurations. The arrow (→) indicates downsampling from fine (simulation) to coarse (training) grids.} 
  \label{tab:datasets}
  \vspace{-6pt}
  \begin{tabular}{lccccccc}
    \toprule
    Dataset & Spatial Grid & Time Steps & \# Training Traj. & \# Testing Traj. & Sample Length & Simulation Duration (s) \\
    \midrule
    Burgers' Equation & 100$^2$→25$^2$ & 1800→450 & 5 & 10 & 20 & 4.5 \\
    Decaying Flow & 2048$^2$→64$^2$ & 76800→2400 & 10 & 10 & 32 & 16.8 \\
    Forced Flow & 2048$^2$→64$^2$ & 38400→1200 & 10 & 10 & 32 & 8.4 \\
    Shear Flow & 2048×1024→128×64 & 44800→1400 & 5 & 10 & 20 & 14.0 \\
    \bottomrule
  \end{tabular}
\end{table*}

Given a solution $\mathbf{u}_k$ at time step $k$, the model predicts the solution $\mathbf{u}_{k+1}$ at the next time step through:
\begin{equation}
\mathbf{u}_{k+1} = \mathbf{u}_k + \int_{t_k}^{t_{k+1}} H(\mathbf{u_k}, \tau) \, d\tau,
\end{equation}
where $H(\mathbf{u_k}, \tau)$ represents the FV Module, a learnable operator approximating the residual of the governing equation. Temporal integration is performed using the fourth-order Runge-Kutta (RK4) method, achieving $O(\Delta t^4)$ accuracy with $\Delta t = t_{k+1} - t_k$ (implementation details are provided in Appendix~\ref{appendix:rk4}). Large $\Delta t$ values on coarse grids exacerbate temporal error accumulation, motivating our analysis and the development of the temporal correction block. The temporal error is defined as:
\begin{equation}
e_k = \mathbf{u}_k - \mathbf{u}_k^{\dagger},
\end{equation}
where $\mathbf{u}_k$ denotes the coarse-grid solution and $\mathbf{u}_k^{\dagger}$ the exact solution. Appendix~\ref{appendix:errro-propagation} derives the propagation relation:
\begin{equation}
e_k = g^{(k-1)}(\mathbf{u}_0, \dots, \mathbf{u}_{k-1}, \Delta t, e_0),
\end{equation}
with $g^{(k-1)}$ representing the error propagation function after $k-1$ iterations. This formulation establishes $e_k$'s dependence on prior numerical solutions $\mathbf{u}_0, \dots, \mathbf{u}_{k-1}$ and their cumulative errors. Our neural correction mechanism processes these solutions through a FNO model, chosen for its ability to efficiently capture global features and handle multiscale resolution challenges::
\begin{equation}
e_k = Q\left(\mathcal{F}^{-1}\left(R_L \cdot \mathcal{F}(\mathbf{u}_0, \dots, \mathbf{u}_{k-1})\right)\right).
\end{equation}
Considering the complexity of subsequent experiments, this module is exclusively implemented in the forced flow. The hyperparameters are provided in Appendix Table~\ref{tab:network_hyperparameters}.

\subsection{Poisson Solver and Projection process}

In solving the NSE for incompressible flows, the pressure field is derived by solving a Poisson equation via a projection method (see Appendix Section~\ref{appendix:projection}). The Poisson equation is given by $\Delta p = \mathbf{\nabla} \cdot \mathbf{u}_k^*$, where $\mathbf{u^*}$ is the intermediate velocity field. To estimate the pressure field, we employ a spectral method, with details provided in Appendix Section~\ref{appendix:projection}. Based on $\mathbf{\nabla} \cdot \textbf{u}^* $, we update the pressure $p^k$ (see Figure \ref{fig:network}(b)). Then, we obtain the next step of divergence-free $\mathbf{u}_{k+1}$ based on the projection step, $\mathbf{u}_{k+1} = \mathbf{u}_k^* - \Delta t \cdot \mathbf{\nabla} p$.

\section{Experiments}
We evaluated the performance of LDSolver on several benchmark flow problems, including Burgers, decaying, forced, and shear flow. The results show that our model achieves higher accuracy and better generalizability, thanks to its unique design that fully integrates with a traditional solver. We have released the source code and data at \href{https://github.com/intell-sci-comput/LDSolver.git}{https://github.com/intell-sci-comput/LDSolver.git}.

\begin{figure*}[t!]
  \centering
  \includegraphics[width=0.97\linewidth]{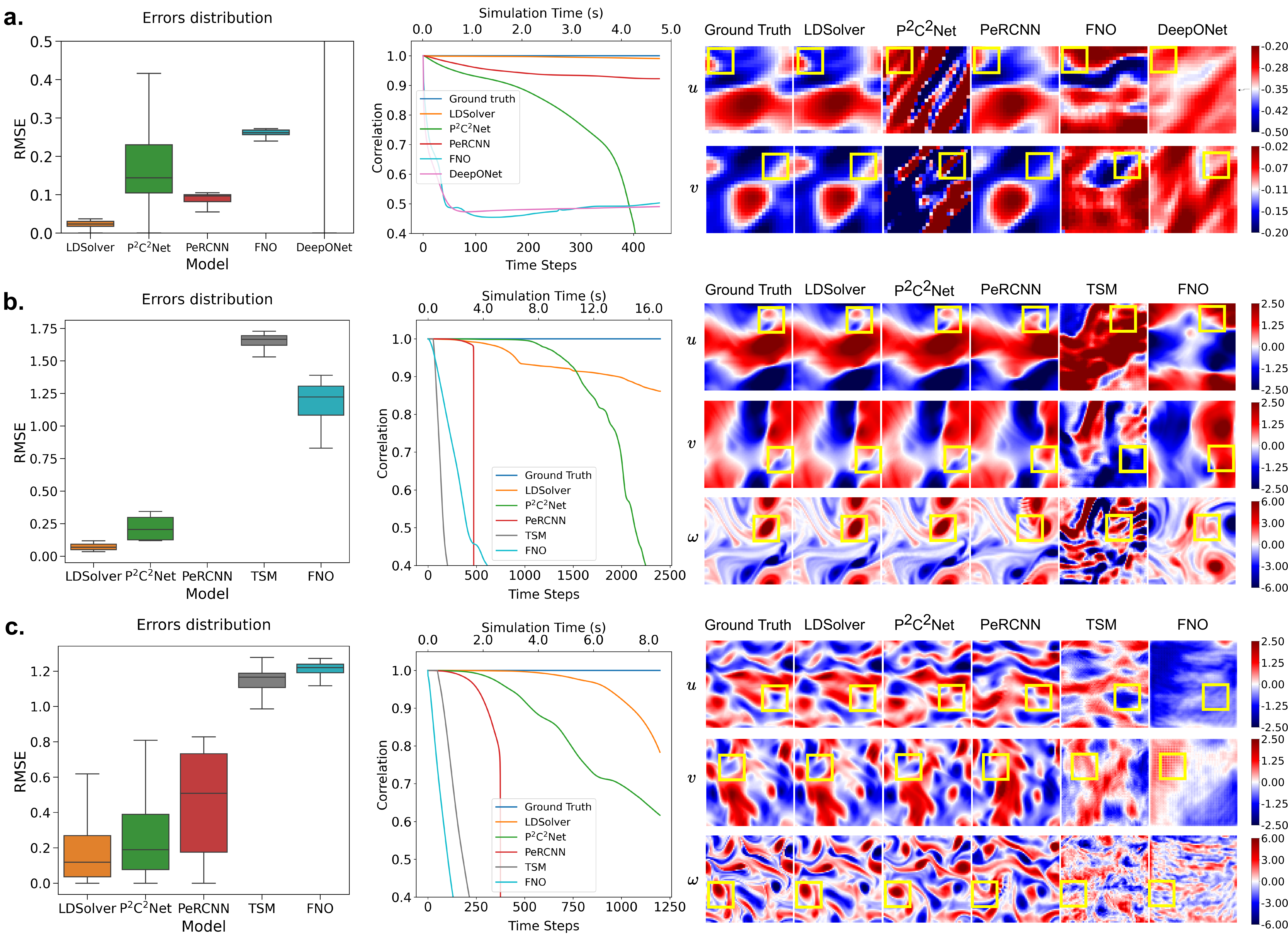}
  \caption{Comparison between LDSolver and baseline models, showing error distributions (left), correlation coefficient time series (middle), and predicted solutions at final time steps (right). (a) Burgers' Equation ($25\times25$); (b) Decaying Flow ($64\times64$); (c) Forced Flow ($64\times64$). Snapshots on the right represent the final simulation states at 4.5\,s, 16.8\,s, and 8.4\,s, respectively.}
  \Description{Visual comparison of LDSolver and baselines including error maps, correlation coefficient curves over time, and final predicted states for Burgers' Equation, Decaying Flow, and Forced Flow on different grid sizes.}
  \label{fig:main_reuslt1}
\end{figure*}

\begin{figure*}[t!]
  \centering
  \includegraphics[width=0.92\linewidth]{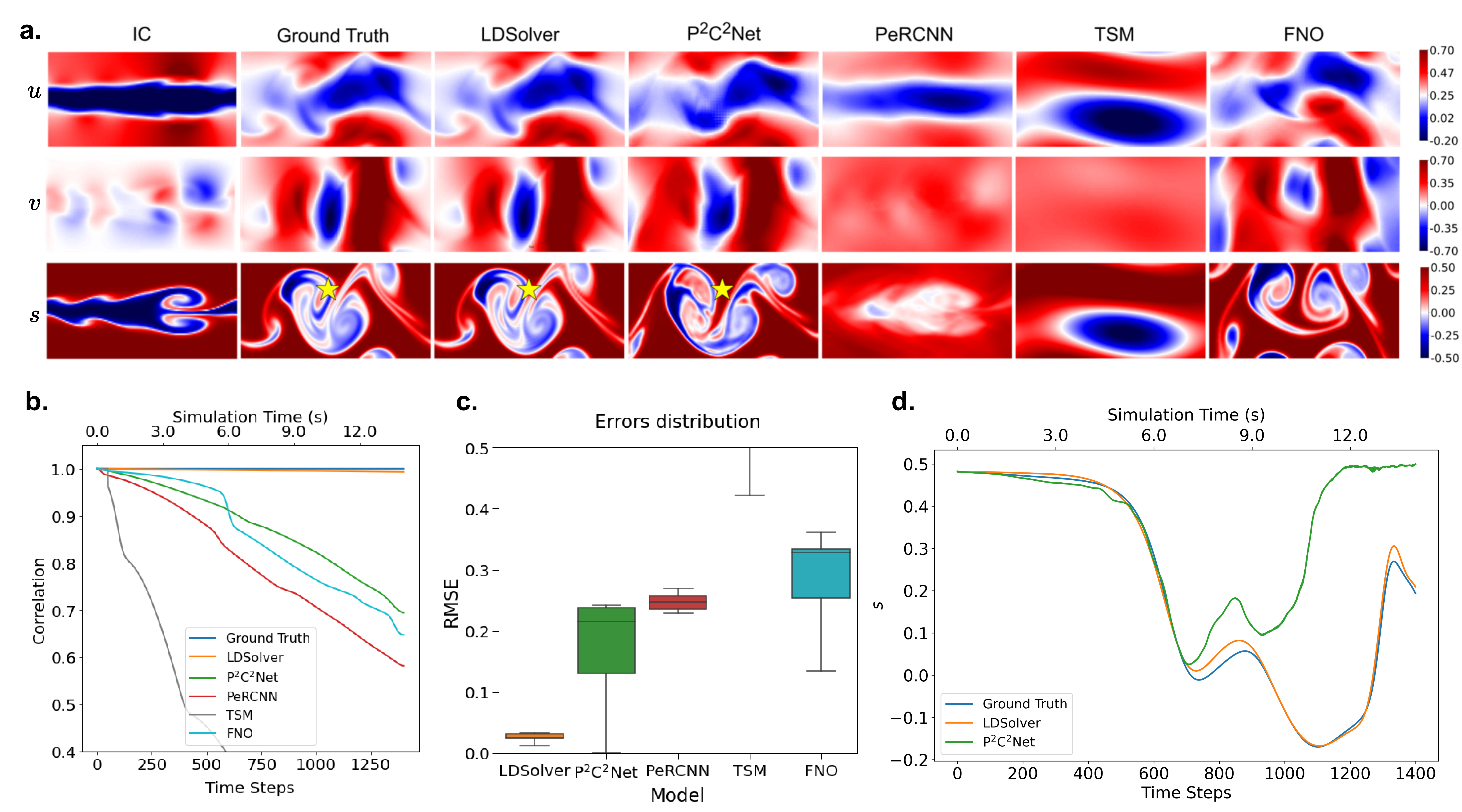}
  \caption{Comparison between LDSolver and baseline models for predicted solutions of shear flow. The models are trained on a grid size of $64 \times 128$. (a) The first column shows the initial conditions (ICs), with the remaining columns displaying snapshots at the final simulation state $t = 14$ s for different models. The sampling point is marked by a star (\textcolor{yellow}{$\bigstar$}) at coordinates $(48, 64)$. (b) Correlation time series of different baseline models. (c) Error distribution. (d) Time series of the variable $s$ in shear flow at the star (\textcolor{yellow}{$\bigstar$}) sampling point for the ground truth, LDSolver, and P$^2$C$^2$Net.}
  \Description{Results for shear flow prediction showing initial conditions, final state snapshots at t = 14s, correlation over time, error maps, and signal evolution at a fixed sampling point for various models including LDSolver.}
  \label{fig:main_reuslt2}
\end{figure*}

\subsection{Experimental Setup} 

\textbf{Simulation Datasets.} We consider four types of two-dimensional flows: Burgers's equation, decaying flow, forced flow, and shear flow. These distinct flow types present diverse challenges for LDSolver. By activating only the physics-based convolution, our model reduces to a traditional numerical solver. Simulations on fine spatio-temporal grids yield high-fidelity flows, denoted as $\tilde{\textbf{u}} \in \mathbb{R}^{\tilde{T} \times n \times \tilde{n}_y \times \tilde{n}_x}$. Coarse-grid flows, represented as $\textbf{u} \in \mathbb{R}^{T \times n \times n_y \times n_x}$, are obtained by downsampling the fine-grid fields in space and time. These coarse-grid flows serve as training labels. For each flow type, we generate 5–10 trajectories for training and evaluate performance on 10 trajectories. Dataset parameters are detailed in Table \ref{tab:datasets}, with additional specifications in Appendix Table \ref{tab:dataset details}.  

\textbf{Model Training.} We partition the data along the temporal dimension to make samples. Each sample has a length of $k_s$, resulting in $n_{\text{sample}} = N / k_s$ samples. The model roll out in an autoregressive manner to get predictions. The loss is computed at the predicted flow field $\textbf{u}$ with the corresponding labels $\check{\textbf{u}}$. The loss function is:
\begin{equation}
\mathcal{L}(\boldsymbol{{\theta}}) = \frac{1}{MB} \sum_{i=1}^{M} \sum_{j=1}^{B} \text{MSE}(\check{\textbf{u}}_{ij}, \textbf{u}_{ij}),    
\end{equation}
where $\mathbf{u}_{ij}$ represents the predicted flows of the $j$-th sample in the $i$-th batch, $\check{\textbf{u}}_{ij}$ is the corresponding label, $M$ denotes the number of batches, $B$ is the batch size, and $\boldsymbol{\theta}$ represents the trainable parameters. Detailed network parameters and training configurations are provided in Appendix Section~\ref{ap:training details}.

\textbf{Evaluation Metrics.} The model’s performance is evaluated using four key metrics: Root Mean Squared Error (RMSE) and Mean Absolute Error (MAE) quantify the average magnitude of errors; Mean Normalized Absolute Difference (MNAD) assesses the relative error; and High Correlation Time (HCT) measures the time required for the model to achieve high-fidelity accuracy. Formal definitions of these metrics can be found in Appendix Section \ref{appendix:metrics}.

\textbf{Baseline Models.} To demonstrate the superiority of the proposed LDSover, we conduct comparisons with several benchmark models, including FNO~\cite{LiKovachkiAzizzadenesheliLiuBhattacharyaStuartAnandkumar2021}, PeRCNN~\cite{Rao2023}, DeepONet~\cite{Lu2018}, TSM~\cite{sun2023neural} and P$^2$C$^2$Net~\cite{wang2024p}. Detailed descriptions and training configurations of these benchmark models are provided in Appendix Section~\ref{appendix:baseline models}. Note that we ignored to select  Learned Interpolation (LI)~\cite{Kochkov2021} as the baseline because it has the instability issue that leads to NaN values in prediction, particularly when the training data is limited and sparse~\cite{wang2024p}. We compared LDSolver with these baseline models across various metrics in Appendix~\ref{ap:baseline compare}, demonstrating that LDSolver exhibits highly general capabilities.

\subsection{Main Results}

Figures~\ref{fig:main_reuslt1} and \ref{fig:main_reuslt2} show the results comparison between LDSolver and baseline models, including error distributions, correlation coefficient time series, and snapshots at the final time step. The quantitative performance results are shown in Table~\ref{table:performance-comparison}.

\textbf{Burgers' Equation.} The Burgers' equation involves advection-diffusion processes. LDSolver, PeRCNN, and FNO successfully capture the dynamics, while P2C2Net suffers from poor physical stability due to lack of local conservation. However, our method demonstrates a significant advantage in error levels, as evidenced by the error distributions in Figure~\ref{fig:main_reuslt1}(a) (left). Table~\ref{table:performance-comparison} further supports this observation, showing substantial performance gains of our model over the baselines, with improvements exceeding 70\% in spatial metrics during long-term predictions.

\textbf{Decaying Flow.} We evaluated LDSolver on a decaying flow at a Reynolds number of $Re = 1000$ with no external force. As shown in Figure~\ref{fig:main_reuslt1}(b) (right), all baseline models except TSM struggled to accurately predict the trajectory. In contrast, LDSolver demonstrated superior stability and effectively learned the underlying dynamics. This is further supported by the error analysis in Figure~\ref{fig:main_reuslt1}(b) (left), where LDSolver achieved the lowest error compared to the baselines. Table~\ref{table:performance-comparison} provides a comprehensive summary of performance metrics, showing improvements exceeding 55\% error reduction for LDSolver across all evaluations in the decaying flow scenario.

\begin{figure}[t!]
\vskip 0in
\begin{center}
\centerline{\includegraphics[width=0.44\textwidth]{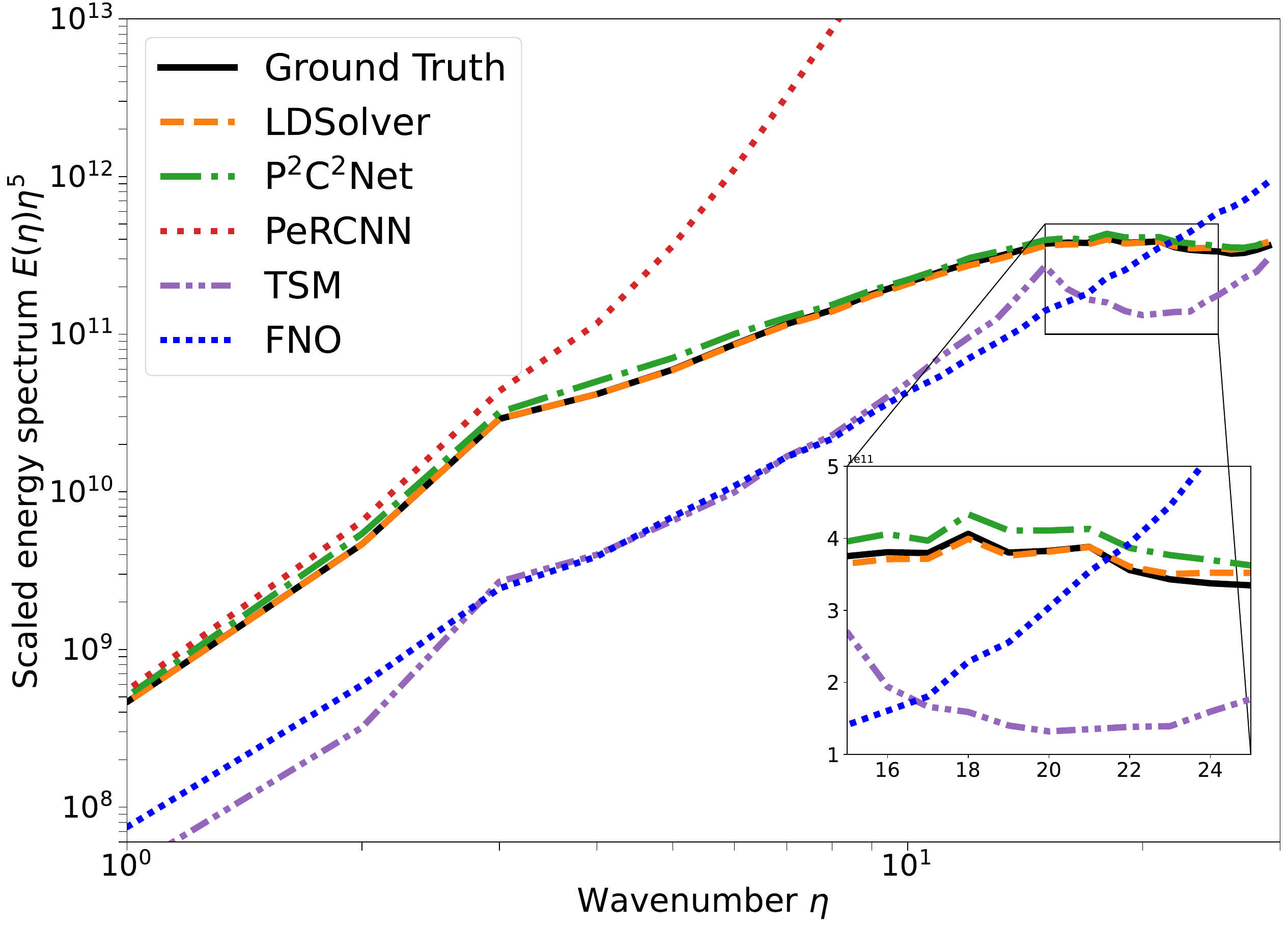}}
\caption{Energy spectra of different baseline models for the forced flow.}
\Description{A line plot comparing the energy spectra of various baseline models applied to the forced flow case.}
\label{fig:spectra}
\end{center}
\vspace{-6pt}
\end{figure}

\textbf{Forced Flow}. We further evaluated LDSolver on a forced flow with external forcing at $Re = 1000$, a benchmark dataset known for its significant challenges~\cite{Kochkov2021,Sun2023}. As shown in Figure~\ref{fig:main_reuslt1}(c), snapshots generated by baseline models at $t = 8.4$ seconds exhibited incorrect dynamic patterns. Our model achieved the lowest average test error compared to the baselines. Additionally, LDSolver consistently outperformed the baselines in terms of correlation coefficient evolution. Table~\ref{table:performance-comparison} further confirms the superior performance of our model, demonstrating improvements of at least 44\% across all metrics. We also analyzed the learned physical properties of the fluid dynamics, such as the energy spectrum. As illustrated in Figure~\ref{fig:spectra}, the energy spectrum of LDSolver shows excellent agreement with the ground truth, highlighting its capability to capture high-frequency features. Furthermore, we validated the results of forced flow by comparing snapshots and correlations between DNS at different resolutions and LDSolver. As shown in Appendix Figure~\ref{fig:compare DNS snapshot} and Appendix Figure~\ref{fig:compare DNS corrolation}, LDSolver's accuracy surpasses that of the 512$\times$512-resolution DNS and approaches that of the 1024$\times$1024-resolution DNS. This enables significant simulation acceleration.

\textbf{Shear Flow.} We further evaluated LDSolver on the NSE with mixture transport~\cite{burns2020dedalus}. This flow requires solving a coupled advection-diffusion equation alongside the NSE, which increases the learning complexity of the model. As illustrated in the predicted snapshots in Figure~\ref{fig:main_reuslt2}(a), few baseline models captured the global patterns but exhibited limitations in resolving local details. In contrast, LDSolver demonstrated a significant advantage in learning scalar transport dynamics. Moreover, the error analysis in Figure~\ref{fig:main_reuslt2}(c) revealed that LDSolver achieved errors nearly 1 to 2 orders of magnitude lower than the baselines. Table~\ref{table:performance-comparison} highlights performance improvements ranging from 60\% to 87\% compared to the best-performing baseline. Additionally, the time series at the sampling point in Figure~\ref{fig:main_reuslt2}(d) also shows that LDSolver closely aligns with the ground truth.

\begin{table}[t!]
\caption{Benchmark comparisons of LDSolver against baseline methods. Performance promotion (PROM) indicates the relative metric change ratio between LDSolver and the top-performing baseline.}
\label{table:performance-comparison}
\vspace{-6pt}
\begin{center}
\begin{small}
\begin{sc}
\resizebox{0.49\textwidth}{!}{\begin{tabular}{lccccc}
\toprule
\textbf{Case} & \textbf{Model} & \textbf{RMSE} & \textbf{MAE} & \textbf{MNAD} & \textbf{HCT (s)} \\
\midrule
\multirow{6}{*}{Burgers' Equation}
&DeepONet&4228.2& 9107.8&6443.1&0.1000\\
&FNO&0.2370 &0.1951 &0.1627&0.2000\\
&PeRCNN&\underline{0.0757}& \underline{0.0543}& \underline{0.0422}&\underline{4.5000}\\
&P$^2$C$^2$Net &0.1357 &0.0907 &0.0781&3.0000\\
&\textbf{LDSolver}& \textbf{0.0220} & \textbf{0.0145} & \textbf{0.0111} & \textbf{4.5000} \\
\cmidrule{2-6}
& Prom. (↑) &  \textcolor{blue}{70.91\%} & \textcolor{blue}{73.26\%} & \textcolor{blue}{73.70\%} & \textcolor{blue}{0\%} \\
\midrule
\multirow{6}{*}{Decaying Flow}
&FNO&1.1669 &0.9055 &0.1282&2.5749\\
&TSM&1.6478& 1.2901&0.1851&0.2103\\
&PeRCNN&nan& nan& nan&3.4704\\
&P$^2$C$^2$Net &\underline{0.2184} &\underline{0.0931} &\underline{0.0130}&\underline{12.6200}\\
&\textbf{LDSolver}& \textbf{0.0732} & \textbf{0.0407} & \textbf{0.0056} & \textbf{16.8262}  \\
\cmidrule{2-6}& Prom. (↑) & \textcolor{blue}{66.48\%} & \textcolor{blue}{56.28\%} & \textcolor{blue}{56.92\%} & \textcolor{blue}{33.33\%} \\
\midrule
\multirow{6}{*}{Forced Flow}
&FNO&1.1684 &0.9180 &0.1166&0.4908\\
&TSM&1.1235 &0.8729 &0.1106&0.5609\\
&PeRCNN&nan& nan& nan&2.6291\\
&P$^2$C$^2$Net &\underline{0.4298} &\underline{0.2690} &\underline{0.0322}&\underline{5.2582}\\
&\textbf{LDSolver}& \textbf{0.2377} & \textbf{0.1253} & \textbf{0.0159} & \textbf{8.4131} \\
\cmidrule{2-6}
& Prom. (↑) & \textcolor{blue}{44.70\%} & \textcolor{blue}{53.42\%} & \textcolor{blue}{50.62\%} & \textcolor{blue}{60.00\%} \\
\midrule
\multirow{6}{*}{Shear Flow}
&FNO&0.1847 &0.0996 &0.0997&8.6000\\
&TSM&0.4337 &0.3424 &0.3427&1.5000\\
&PeRCNN&0.2470 &0.1985 &0.1987&7.0000\\
&P$^2$C$^2$Net&\underline{0.1834} &\underline{0.0981} &\underline{0.0982}&\underline{8.7000}\\
&\textbf{LDSolver}&\textbf{0.0231} &\textbf{0.0129} &\textbf{0.0129}&\textbf{14.0000}\\
\cmidrule{2-6}
& Prom. (↑) & \textcolor{blue}{87.40\%} & \textcolor{blue}{86.85\%} & \textcolor{blue}{86.86\%} & \textcolor{blue}{60.92\%} \\
\bottomrule
\end{tabular} }
\end{sc}
\end{small}
\end{center}
\end{table}

\subsection{Generation Tests}

\begin{figure}[t!]
\vskip 0in
\begin{center}
\centerline{\includegraphics[width=0.48\textwidth]{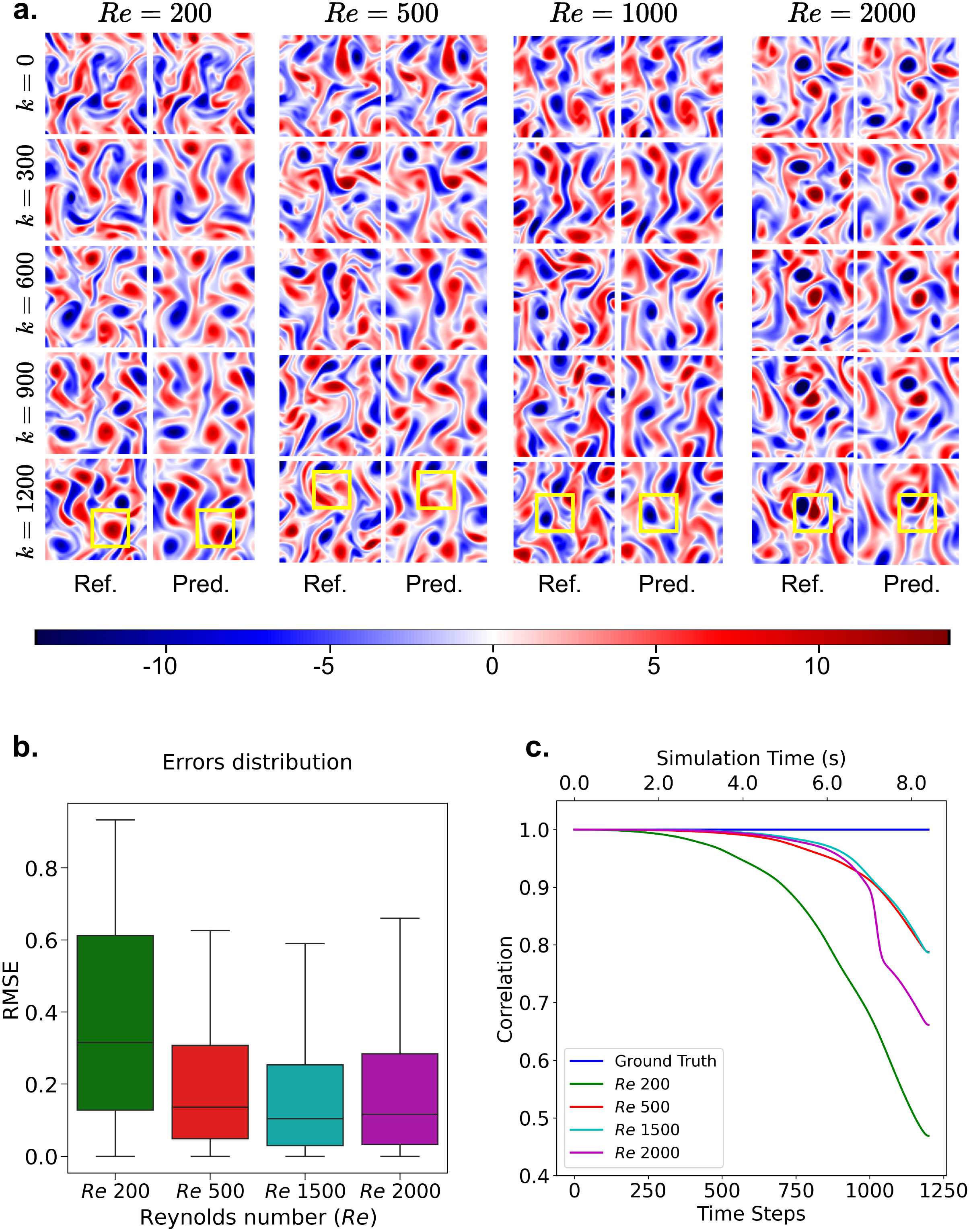}}
\caption{LDSolver generalization over various Reynolds numbers ($Re$). (a) Vorticity evolution snapshots for forced flow. (b) Error distribution. (c) Correlation time series.}
\Description{Visualization of LDSolver's generalization to different Reynolds numbers, including vorticity snapshots of forced flow, corresponding error maps, and correlation curves over time.}
\label{fig:generation}
\end{center}
\vspace{-6pt}
\end{figure}

Using the forced flow and shear flow case, we evaluated the generalization capability of LDSolver across different PDE parameters. The model was trained at $Re = 1000$ and tested at four Reynolds numbers: $Re = \{200, 500, 1500, 2000\}$. As shown in Figure~\ref{fig:generation}(b), the error distributions demonstrated stable performance, consistently remaining below a threshold across all tested Reynolds numbers. Figure~\ref{fig:generation}(c) further confirms robust and satisfactory error levels, with smaller errors observed when the Reynolds number is closer to the training value, as validated by the correlation curves in Figure~\ref{fig:generation}(b). We observed a notable poor performance at $Re = 200$, which we hypothesize is due to the exponential dependence of the ratio between the nonlinear convective term and the linear viscous term in the NSE on the Reynolds number ($Re$). As $Re$ decreases from 1000 to 200, the dominant term shifts from the convection term $(\mathbf{u} \cdot \nabla) \mathbf{u}$ to the linear dissipationi term $\nu \nabla^2 \mathbf{u}$. However, the model parameters remain fixed based on the high-$Re$ feature extraction mode, lacking the ability to adaptively adjust the feature weight distribution. Extended generalization tests on shear flow with varying $Re$ and diffusion coefficients ($D$) are provided in Appendix~\ref{ap:additional results}.

\subsection{Ablation Study}

To assess the contributions of individual components to model performance, we performed an ablation study on the forced flow dataset using six configurations: (1) Model 1 (physics-based convolution only, equivalent to DNS solver); (2) Model 2 (without physics-based convolution); (3) Model 3 (without frequency domain operator); (4) Model 4 (frequency domain operator exclusively); (5) Model 5 (without Temporal Correction block); and (6) the complete LDSolver architecture. All experiments maintained identical dataset and hyperparameter settings as specified in previous sections.

\begin{table}[t!]
\caption{Ablation study results for LDSolver.}
\label{tab:ablation}
\vspace{-6pt}
\begin{center}
\begin{tabular}{lcccc}
\toprule
Model & RMSE & MAE & MNAD & HCT (s) \\
\midrule
DNS Solver & 0.8333  &0.6572  &0.1313  &3.5475  \\
w/o PhyConv & 0.2874 &0.2155 & 0.0386 & 6.9478 \\
w/o FreqOp & NAN& NAN & NAN & 1.5914 \\
FreqOp only & NAN& NAN& NAN& 1.8158\\
w/o TempCorr & 0.3361 & 0.2542 & 0.0456 & 6.5622  \\
\textbf{LDSolver (Full)} & \textcolor{blue}{0.2377} & \textcolor{blue}{0.1253} & \textcolor{blue}{0.0159} & \textcolor{blue}{8.4131} \\
\bottomrule
\end{tabular}
\end{center}
\vspace{-6pt}
\end{table}

Table~\ref{tab:ablation} demonstrates the significant impact of each component. The findings are presented as folows. (1) With only the physics-based convolution, the model becomes untrainable, effectively reducing LDSolver to a traditional solver. Consequently, direct solutions on coarse grids using this configuration result in substantial numerical errors, exhibiting error levels four times higher than the complete model. This highlights the critical role of the learnable modules on coarse grids. (2) Removing the physics-based convolution, which provides the baseline solution derived from fixed kernels, leads to degraded performance. (3) The absence of the frequency domain operator introduces convergence issues during inference. (4) Eliminating the temporal correction block significantly increases cumulative errors. In summary, the physics-based convolution provides a basic solution on coarse grids, the learnable convolutions capture local information, frequency domain operator supplys global information and enhance inference stability, and the temporal correction block ensures accurate predictions over long time horizons. These components are therefore essential.

\section{Conclusion}

\begin{table}[t!]
\caption{Computation Speed Comparison. Values represent GPU simulation time required per physical second. The DNS resolutions for the four flow systems are $100\times100$, $1024\times1024$, $1024\times1024$, and $1024\times512$, respectively. The corresponding LDSolver resolutions are $25\times25$, $64\times64$, $64\times64$, and $128\times64$.}
\label{tab:speed}
\vspace{-6pt}
\begin{center}
\begin{tabular}{lcccc}
\toprule
Case & Burgers & Decaying & Forced & Shear \\
\midrule
DNS & 2.3067 & 29.2402 & 32.8061 & 29.1428 \\
LDSolver & 0.9800 & 2.3773 & 3.8036 & 2.7857 \\
\textbf{Speed up $\times$ (↑)} & \textcolor{blue}{2.3537} & \textcolor{blue}{12.3000} & \textcolor{blue}{8.6250} & \textcolor{blue}{10.4615} \\
\bottomrule
\end{tabular}
\end{center}
\vspace{-6pt}
\end{table}

We introduce a learnable and differentiable finite volume solver (LDSolver) for efficient and accurate simulation of fluid flows on coarse spatiotemporal grids. LDSolver consists of two core components: (1) a differentiable finite volume solver as the computational framework, and (2) learnable modules designed to correct derivative, interpolation, and temporal errors on coarse grids. Within the spatial module, a physics-based convolution provides a baseline solution, learnable convolutions extract local features, and a frequency domain operator captures global features. In the temporal module, a temporal correction block mitigates accumulated errors in long-term simulations. Experiments on four flow systems demonstrate that LDSolver significantly accelerates simulations (as shown in Table~\ref{tab:speed}) while maintaining high accuracy and robust generalization, even with limited training data. While LDSolver shows great potential, two key areas require further exploration: (1) its reliance on regular grids limits its applicability to complex geometries. This limitation can be addressed by incorporating geometry-aware models (e.g., Geo-FNO~\cite{li2023fourier}), coordinate transformations (e.g., PhyGeoNet~\cite{gao2021phygeonet}), or graph-based extensions (e.g., PhyMPGN~\cite{zeng2025phympgn}) to handle irregular domains and boundary conditions, and (2) extending the current framework to support three-dimensional flows remains a critical direction for future research.

\begin{acks}
This work is supported by the National Natural Science Foundation of China (No. 62276269, No. 92270118) and the Beijing Natural Science Foundation (No. 1232009).
\end{acks}

\bibliographystyle{ACM-Reference-Format}
\bibliography{reference}


\clearpage
\appendix

\renewcommand{\thefigure}{S\arabic{figure}}
\setcounter{figure}{0} 

\renewcommand{\theequation}{S\arabic{equation}}
\setcounter{equation}{0} 

\renewcommand{\thetable}{S\arabic{table}}
\setcounter{table}{0} 


\begin{center}
    {\Large \textbf{Appendix}}
\end{center}

This supplementary material file provides the appendix section to the main article.

\section{Notation}

The commonly used symbols and variables throughout the article are listed in Table~\ref{tab:notation}.

\begin{table*}[t!]
\caption{Commonly used symbols and variables and their meaning.}
\label{tab:notation}
\vspace{-6pt}
\begin{center}
\begin{small}
\begin{tabular}{lcccccccc}
\toprule
Variable & Meaning \\
\midrule
 $\textbf{u}_k^{ij}$ & Velocity at 2D grid coordinate $(i,j)$ at time step $k$ on coarse grids \\
    $\tilde{\textbf{u}}_k^{ij}$ & Velocity at 2D grid coordinate $(i,j)$ at time step $k$ on fine grids \\
    $\bar{\mathbf{u}}$ & Average value of $\mathbf{u}$ \\
     $\mathbf{u^*}$ & The intermediate velocity field \\
    $p_{k}^{ij}$ & Pressure at 2D grid coordinate $(i,j)$ at time step $k$ \\
    $\textbf{f}$ & External force \\
    $Re$ & Reynolds number \\
    $V^{ij}$ & Control volume at 2D grid coordinate $(i,j)$ \\
    $A^{ij}$ & Area of the face $(i,j)$ \\
    $k_c$ & Temporal buffer correction steps \\
    $N$ & Total simulation time steps \\
    $T$ & Total simulation time \\
    $\textbf{x}$ & The coarse-grid coordinates \\
    $\delta t$ & The fine time step \\
    $\Delta t$ & The coarse time step \\
    $\delta x$ & The fine grids distance \\
    $\Delta x$ & The coarse grids distance \\
    $k_s$ & Sample length \\
\bottomrule
\end{tabular}
\end{small}
\end{center}
\vskip 0in
\end{table*}

\section{Basics of the Finite Volume Method}
\label{appendix:Discretization}

\subsection{Staggered Grid and BC Encoding}

\begin{figure*}[h!]
\begin{center}
\centerline{\includegraphics[width=0.7\textwidth]{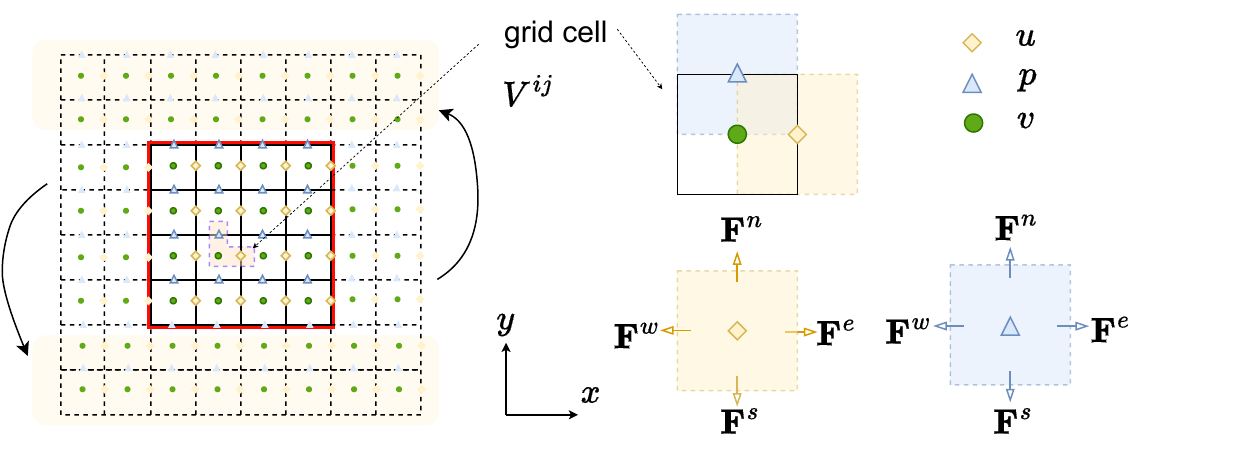}}
\caption{Staggered grid and boundary condition encoding.}
\Description{Illustration of a staggered grid layout and the encoding of boundary conditions, possibly showing velocity and pressure locations on the mesh.}
\label{fig:grid}
\end{center}
\end{figure*}

The staggered grid is an efficient discretization strategy in finite-volume methods, where velocity components and pressure (or other scalar fields) are stored at distinct spatial positions within grid cells, as illustrated in Figure~\ref{fig:grid}. For a control volume $V^{ij}$, velocity components are typically located at cell face centers, while pressure resides at cell centroids. This staggered arrangement inherently avoids non-physical oscillations (e.g., checkerboard artifacts) caused by pressure-velocity decoupling in traditional collocated grids, leveraging natural interpolation of physical variables. Within the finite-volume framework, staggered grids enable precise flux integration of mass and momentum across cell boundaries by directly utilizing velocity values at face locations, ensuring strict local conservation. Furthermore, pressure gradient terms are naturally computed through finite differences of adjacent pressure values, eliminating interpolation errors and enhancing numerical stability.

To rigorously enforce prescribed periodic boundary conditions and maintain feature map consistency during differentiation, our architecture integrates periodic boundary condition (BC) padding (Figure~\ref{fig:grid}). This deterministic padding approach ensures precise boundary periodicity while enhancing numerical accuracy in derivative-based computations.

\subsection{Spatial Discretization}

Taking the $x$-direction as an example, we consider the discretization of the momentum equation (Eq.~\ref{eq:momontum_eq}) using the finite volume method (FVM). We denote each term in the NSE below, where all the discretizations are performed within a control volume $V$ (with the indices $i, j$ omitted for simplicity). Using the divergence theorem, we transform the volume integrals into surface integrals, leading to eq.~\ref{eq:transient_term}:
\begin{equation}
\begin{aligned}
\underbrace{\frac{\partial}{\partial t} \int_V u \, dV}_{\text{(Transient term)}} = &
-\underbrace{ \oint_S u \mathbf{u} \cdot \mathbf{n} \, dS}_{\text{(Advection term)}} 
\ \ - \underbrace{ \oint_S p \mathbf{n} \, dS}_{\text{(Pressure term)}} \\
& + \ \underbrace{\frac{1}{Re} \oint_S \boldsymbol{\boldsymbol{\nabla}} u \cdot \mathbf{n} \, dS}_{\text{(dissipation term)}} \ \ \ \ \  
 + \underbrace{\oint_S f_x \, dS}_{\text{(External force term)}}.
\end{aligned}
\label{eq:transient_term}
\end{equation}
This equation represents the transient term in its integral form.

\textbf{Advection Term:} The second term involves the advection of the velocity field, expressed as \( (\mathbf{u} \cdot \boldsymbol{\boldsymbol{\nabla}}) \mathbf{u} \). By applying the divergence theorem, the volume integral is converted to a surface integral:
\begin{equation}
\oint_S \mathit{u} \mathbf{u} \cdot \mathbf{n} \, dS \approx (uu)_e \Delta y - (uu)_w \Delta y + (vu)_n \Delta x - (vu)_s \Delta x.
\label{eq:advection_term}
\end{equation}
Here, \(e,w,n,s\) denote the east, west, north, and south faces of the control volume, respectively.

\textbf{Dissipation Term:} The dissipation term represents the viscous forces in the fluid and is given by \( \frac{1}{Re} \boldsymbol{\boldsymbol{\nabla}}^2 \mathbf{u} \). Using the divergence theorem again, we transform the volume integral into surface integrals:
\begin{equation}
\oint_S \boldsymbol{\nabla} u \cdot \mathbf{n} \, dS = 
\left. \frac{\partial u}{\partial x} \right|_e \Delta y
- \left. \frac{\partial u}{\partial x} \right|_w \Delta y
+ \left. \frac{\partial u}{\partial y} \right|_n \Delta x
- \left. \frac{\partial u}{\partial y} \right|_s \Delta x.
\label{eq:dissipation_term}
\end{equation}

\textbf{Pressure Term:} The pressure term is discretized by applying the Gauss theorem:
\begin{equation}
\oint_S p \mathbf{n} \, dS \approx \left (p_e \Delta y - p_w \Delta y \right) = \left (p_e - p_w) \Delta y \right.
\label{eq:pressure_term}
\end{equation}

\textbf{External Forcing Term:} Finally, the external forcing term \( \mathbf{f} \) is assumed to be constant within each control volume, and its integral is approximated as:
\begin{equation}
\oint_S f_x \, dS \approx {f_x} \cdot V.
\end{equation} 
In this process, evaluating the convective and diffusive fluxes requires estimating velocity interpolations and spatial derivatives at the faces of control volumes within $V$. Using Taylor expansion and linear interpolation, these quantities at the cell faces can be approximated as:
\begin{equation}
    \label{eq:physics_kernels}
W_P^{(d)} = {
    \begin{bmatrix} 
        0 & 0 & 0 & 0  \\
        0 & 0 & 0 & 0  \\
        1 & -27 & 27 & -1  \\
        0 & 0 & 0 & 0  \\  
        0 & 0 & 0 & 0  
    \end{bmatrix}} \times \frac{1}{24(\Delta x)},
\end{equation}
\begin{equation}
W_P^{(i)} = {
    \begin{bmatrix} 
        0 & 0 & 0 & 0  \\
        0 & 0 & 0 & 0  \\
        0 & -1 & 1 & 0  \\
        0 & 0 & 0 & 0  \\  
        0 & 0 & 0 & 0  
    \end{bmatrix}} \times \frac{1}{2(\Delta x)},
\end{equation}
where $W_P^{(d)}$ represents the kernel for derivative estimation, which employs a fourth-order accurate finite difference scheme, and $W_P^{(i)}$ denotes the kernel for interpolation, utilizing a second-order accurate central difference scheme. These kernels are applied to approximate the required quantities at the control volume faces, ensuring accurate numerical solutions.

The pressure term is typically addressed using projection methods, as detailed in Appendix Section \ref{appendix:projection}. For time integration, various schemes can be employed; in this study, we utilize the fourth-order Runge-Kutta (RK4) method, the specifics of which are provided in the Appendix Section~\ref{appendix:rk4}.

\subsection{Projection Method}
\label{appendix:projection}
The projection method is a widely used finite volume approach for solving the NSE, primarily addressing the pressure–velocity coupling problem. Its core concept lies in the time-splitting technique: first, the convection and dissipation terms are solved using the velocity field from the previous time step; then, the pressure term is obtained by solving the Poisson equation to satisfy the divergence-free condition; finally, the velocity field at the next time step is computed through time integration. This method ensures fluid incompressibility by projecting the velocity field and, when combined with appropriate numerical techniques, guarantees both computational efficiency and accuracy. The details are as follows:

Initially, the momentum equation is solved by neglecting the pressure gradient term using the RK4 method (see Appendix Section~\ref{appendix:rk4}). Specifically, an intermediate velocity field \(\mathbf{u}_k^*\) is computed using the following equation:
\begin{equation}
  \frac{\partial \mathbf{u}_k^*}{\partial t}
  = -(\mathbf{u}_k\!\cdot\!\nabla)\,\mathbf{u}_k^*
  + \frac{1}{Re}\,\nabla^2\mathbf{u}_k^*
  + \mathbf{f}\,,
\end{equation}
where \(\mathbf{u}_k^*\) is an intermediate velocity that generally does not satisfy \(\nabla\!\cdot\!\mathbf{u}_k^*=0\). Next, enforce incompressibility by solving for the pressure increment \(p\). Applying \(\nabla\!\cdot\!\mathbf{u}_{k+1}=0\) to the corrector
\(\mathbf{u}_{k+1} = \mathbf{u}_k^* - \delta t\,\nabla p\)
yields
\begin{equation}
  \nabla^2 p \;=\; \frac{1}{\delta t}\,\nabla\!\cdot\!\mathbf{u}_k^*\,,
\end{equation}
where \(p\) is the physical pressure, obtained by solving the1 equation with a Poisson solver as detailed in Appendix~\ref{app:poisson_solver}. Update the velocity to divergence-free form:
\begin{equation}
  \mathbf{u}_{k+1}
  = \mathbf{u}_k^* - \delta t\,\nabla p\,,
\end{equation}
where \(\delta t\) is the time step size. This step corrects the velocity field to ensure adherence to the divergence-free condition. The update of the velocity field is now complete. This procedure guarantees \(\nabla\!\cdot\!\mathbf{u}_{k+1}=0\) at each step, while treating advection and diffusion explicitly for computational efficiency and accuracy.

\subsection{Poisson Solver}  
\label{app:poisson_solver}

The goal of the Poisson solver is to determine the state quantities of a system in a two-dimensional spatial domain using the spectral method for a given Laplacian term. The Laplace equation in two dimensions is:
\begin{equation}
    \Delta p = \psi(\textbf{u}^*).
    \label{eq:poisson_equation}
\end{equation}
Applying Fast Fourier Transform (FFT) on Eq.~\ref{eq:poisson_equation}, we get:
\begin{equation}
-(\eta_x^2 + \eta_y^2)\hat{p} = \hat{\psi(\textbf{u}^*)},
\end{equation}
where $\eta_x$ and $\eta_y$ are the wavenumbers along the $x$ and $y$ axes, respectively, assuming $\eta_x^2 + \eta_y^2 \neq 0$ to avoid division by zero. In the frequency domain, we obtain:
\begin{equation}
\hat{p} = \frac{\hat{\psi(\textbf{u}^*)}}{-(\eta_x^2 + \eta_y^2)}.
\end{equation}
Next, we transform the field from the frequency domain to the spatial domain using inverse Fast Fourier Transform (iFFT):
\begin{equation}
    p = \text{iFFT} \left[  \frac{\hat{\psi(\textbf{u}^*)}}{-(\eta_x^2 + \eta_y^2)} \right].
\end{equation}
By applying this process to $\psi(\textbf{u}^*)$, we can efficiently decouple the pressure field without any labeled data.

\subsection{RK4 scheme}
\label{appendix:rk4}

The RK4 (Runge-Kutta 4th order) method is a widely used numerical integration technique for solving ordinary differential equations (ODEs) and partial differential equations (PDEs). It is commonly employed as a time-stepping solver due to its balance between computational efficiency and accuracy. This method approximates the solution by calculating intermediate slopes at several points within each time step.

The general form of the time-stepping integration for advancing from time step $k$ to $k+1$ can be written as:
\begin{equation}
\mathbf{u}_{k+1} = \mathbf{u}_k + \int_{k}^{k+1} H(\mathbf{u}_k (\mathbf{x}, \tau)) d\tau,
\end{equation}
where $\mathbf{u}_{k+1}$ and $\mathbf{u}_k$ represent the solutions at times $k+1$ and $k$, respectively.

RK4 is a high-order integration scheme. It divides the time interval into several small, equally spaced steps to approximate the integral. The update process for the state at each time step can be described as follows:
\begin{equation}
r_1 = H(\mathbf{u}_k, k),
\end{equation}
\begin{equation}
r_2 = H\left(\mathbf{u}_k + \frac{\Delta_t}{2} \times r_1, k + \frac{\Delta_t}{2}\right),
\end{equation}
\begin{equation}
r_3 = H\left(\mathbf{u}_k + \frac{\Delta_t}{2} \times r_2, k + \frac{\Delta_t}{2}\right),
\end{equation}
\begin{equation}
r_4 = H\left(\mathbf{u}_k + \Delta_t \times r_3, k + \Delta_t\right),
\end{equation}
\begin{equation}
\mathbf{u}_{k+1} = \mathbf{u}_k + \frac{1}{6} \Delta_t \left(r_1 + 2r_2 + 2r_3 + r_4\right).
\end{equation}
This approach ensures that the solution is advanced accurately while maintaining a reasonable computational cost. The weighted sum of the intermediate slopes \( r_1, r_2, r_3, \) and \( r_4 \) provides a high-order approximation to the integral.

\subsection{Temporal Error Analysis}
\label{appendix:errro-propagation}

Consider a numerical time-stepping scheme where $\mathbf{u}_k$ denotes the computed solution at step $k$, $\mathbf{u}_k^\dagger$ the exact solution, and $\delta t$ the time step size. The global error is defined as $e_k = \mathbf{u}_k - \mathbf{u}_k^\dagger$. Assume the numerical method follows the recurrence relation $\mathbf{u}_k = \mathcal{F}(\mathbf{u}_{k-1}, \delta t)$, while the exact solution satisfies $\mathbf{u}_k^\dagger = \mathcal{F}(\mathbf{u}_{k-1}^\dagger, \delta t) + \tau(\mathbf{u}_{k-1}^\dagger, \delta t)$, where $\tau$ is the local truncation error.

The error propagation is derived by subtracting these relationships:
\begin{equation}
    e_k = \mathcal{F}(\mathbf{u}_{k-1}, \delta t) - \mathcal{F}(\mathbf{u}_{k-1}^\dagger, \delta t) - \tau(\mathbf{u}_{k-1}^\dagger, \delta t).
\end{equation}
Linearizing $\mathcal{F}$ about $\mathbf{u}_{k-1}^\dagger$ via Taylor expansion yields:
\begin{equation}
    \mathcal{F}(\mathbf{u}_{k-1}, \delta t) \approx \mathcal{F}(\mathbf{u}_{k-1}^\dagger, \delta t) + \frac{\partial \mathcal{F}}{\partial \mathbf{u}}\bigg|_{\mathbf{u}_{k-1}^\dagger} e_{k-1} + \mathcal{O}(\|e_{k-1}\|^2).
\end{equation}
Substituting this into the error equation reveals the recursive structure:
\begin{equation}
    e_k \approx \frac{\partial \mathcal{F}}{\partial \mathbf{u}}\bigg|_{\mathbf{u}_{k-1}^\dagger} e_{k-1} - \tau(\mathbf{u}_{k-1}^\dagger, \delta t) + \mathcal{O}(\|e_{k-1}\|^2),
\end{equation}
showing that errors propagate through both linear amplification of prior discrepancies and accumulation of truncation errors.

Unrolling the recursion through $k$ steps gives:
\begin{equation}
    e_k = \prod_{i=1}^k \frac{\partial \mathcal{F}}{\partial \mathbf{u}}\bigg|_{\mathbf{u}_{i-1}^\dagger} e_0 - \sum_{j=1}^k \prod_{i=j+1}^k \frac{\partial \mathcal{F}}{\partial \mathbf{u}}\bigg|_{\mathbf{u}_{i-1}^\dagger} \tau(\mathbf{u}_{j-1}^\dagger, \delta t) + \mathcal{O}\left(\sum_{m=1}^k \|e_{m-1}\|^2\right).
\end{equation}
This decomposition highlights three key dependencies: (1) Amplification factors tied to the method's Jacobian matrices, which depend on historical solutions through $\mathbf{u}_i^\dagger = \mathbf{u}_i - e_i$; (2) Cumulative truncation errors modulated by preceding Jacobians; (3) Nonlinear coupling effects when $\mathcal{F}$ is nonlinear.

This proves $e_k = g^{(k-1)}(\mathbf{u}_0,...,\mathbf{u}_{k-1}, \delta t, e_0)$ under the conditions that $\mathcal{F}$ is continuously differentiable and errors remain small ($\|e_k\| \ll 1$). The function $g^{(k-1)}$ encapsulates both linear error propagation and nonlinear state dependencies arising from solution-history-dependent Jacobians. This motivates the design of temporal correction block that leverage historical solution buffers to counteract error accumulation.

\section{Numerical Solver}

\subsection{Datasets Description}

To evaluate the generalizability of LDSolver, we constructed diverse datasets encompassing various complex flow systems, including the Burgers' equation, decaying flow, forced flow, and shear flow. To ensure fairness in testing, we randomly selected 10 seeds with Gaussian random field, providing a diverse set of random initial conditions. Additionally, we introduced a warm-up phase to allow the system to reach a steady state before data collection, thereby mitigating the influence of initial transient effects. Further details can be found in Table~\ref{tab:dataset details}.

\begin{table*}[t!]
\caption{Settings for generating datasets.}
\label{tab:dataset details}
\vspace{-6pt}
\begin{center}
\begin{tabular}{lcccccccc}
\toprule
Parameters / Case & Burgers' Equation & Decaying Flow & Forced Flow & Shear Flow \\
\midrule
Spatial Domain & [0, 1]$^2$ & [0, 2$\pi$]$^2$ & [0, 2$\pi$]$^2$ & [0, 8]$\times$[0,4] \\
Grid           & 100$^2$     & 2048$^2$    & 2048$^2$    & 2048$\times$1024 \\
Training Grid  & 25$^2$     & 64$^2$       & 64$^2$      & 128$\times$64  \\
Simulation dt (s) & $1.00 \times 10^{-3}$ & $2.19 \times 10^{-4}$ & $2.19 \times 10^{-4}$ & $3.125 \times 10^{-4}$ \\
Warmup (s) & 0.5 & 40 & 40 & 12 \\
Training data group & 5 & 10 & 10 & 5 \\
Testing data group & 10 & 10 & 10 & 10 \\
Spatial downsample & 16$\times$ & 1024$\times$ & 1024$\times$ & 512$\times$ \\
Temporal downsample & 10$\times$ & 32$\times$ & 32$\times$ & 32$\times$ \\
\bottomrule
\end{tabular}
\end{center}
\vskip 0in
\end{table*}

\textbf{Burgers' Equation.} The Burgers' Equation is a fundamental equation in fluid dynamics, capturing the interplay between convection and diffusion process. It finds applications across various disciplines, including materials science, applied mathematics, and engineering:
\begin{equation}
\frac{\partial \mathbf{u}}{\partial t} = \nu \nabla^2 \mathbf{u} - \mathbf{u} \cdot \nabla \mathbf{u}, \quad t \in [0, T],
\end{equation}
where $\mathbf{u} = \{u, v\} \in \mathbb{R}^2$ represents the fluid velocity, $\nu$ denotes the viscosity coefficient (set to 0.002), and $\nabla$ signifies the gradient operator. We generate the dataset using our numerical finite volume solver (LDSolver). Periodic boundary conditions are imposed on the spatial domain $x \in [0, 1]$. The data is initially generated on a $100^2$ grid and downsampled to a $25^2$ grid for our training experiments. The simulation uses a time step of $\delta t = 1 \times 10^{-3}$ seconds, with a total simulation time of $T = 4.5$ seconds. In the training phase, we use 5 distinct trajectories, each comprising 450 snapshots with a temporal interval of $\Delta t = 10 \delta t$. For the testing phase, we generate 10 additional trajectories, each also containing 450 snapshots.

\textbf{Decaying \& Forced Flow.} 
The decaying and forced flow is governed by the NSE:
\begin{equation}
\frac{\partial \mathbf{u}}{\partial t} + (\mathbf{u} \cdot \boldsymbol{\nabla}) \mathbf{u} = \frac{1}{Re} \boldsymbol{\nabla}^2 \mathbf{u} - \boldsymbol{\nabla} p + \mathbf{f}, \quad t \in [0, T],
\label{eq:momontum_eq}
\end{equation}
\begin{equation}
\boldsymbol{\nabla} \cdot \mathbf{u} = 0,
\label{eq:divergence_free_eq}
\end{equation}
where $\mathbf{u} = \{u, v\} \in \mathbb{R}^2$ is the fluid velocity vector, $p \in \mathbb{R}$ is the pressure, and $Re$ is the Reynolds number, a dimensionless parameter that characterizes flow regimes such as laminar, turbulent, or transitional. Here, $T$ denotes the total simulation time. The decaying and forced flow is governed by the NSE (see Eq.~\ref{eq:momontum_eq}). To generate the dataset, high-resolution (e.g., $2048 \times 2048$ grid cells) simulations were performed using numerical LDSover on fine grids with a small time step $\delta t$. The resulting data were downsampled to coarser grids (e.g., $64 \times 64$ grid cells) with a significantly larger time step $\Delta t = 32 \delta t$ to establish the ground truth. Initial conditions (ICs) were generated by Gaussian random field. For training, we used only 10 sets of labeled data, each containing 2400 snapshots. For testing, 10 different trajectories were generated. The model's performance was evaluated across 5 Reynolds numbers: $Re = \{200, 500, 1000, 1500, 2000\}$. For decaying flow, $\textbf{f}=0$, while for forced flow, we set $\mathbf{f}_k=[\sin 4y - 0.1u_k, -0.1v_k]^T$.

\textbf{Shear Flow.} Consider a more complex scenario where the NSE are coupled with a Burgers' equation to model the spatiotemporal evolution of a substance (e.g., heat, pollutants, solutes) within a flow. This coupled system finds broad applicability in processes such as atmospheric pollutant diffusion, heat transfer, and solute transport. Here, we take a classic shear flow system as an example and use LDSolver to attempt solving such problems.
\begin{equation}
\frac{\partial \mathbf{u}}{\partial t} + \nabla p - \frac{1}{Re} \nabla^2 \mathbf{u} = - \mathbf{u} \cdot \nabla \mathbf{u}, \quad t \in [0, T].
\end{equation}
This equation describes the fluid motion, where $\mathbf{u} = (u, v) \in \mathbb{R}^2$ is the velocity field, and $p \in \mathbb{R}$ is the pressure, subject to the constraint $\int p = 0$ (i.e., the total pressure is zero).  The velocity field also satisfies the mass conservation equation $\nabla \cdot \mathbf{u} = 0$.
\begin{equation}
\frac{\partial s}{\partial t} - D \nabla^2 s = - \mathbf{u} \cdot \nabla s, \quad t \in [0, T].
\end{equation}
This equation governs the transport of a substance, where $s \in \mathbb{R}$ is a passive scalar representing the concentration or temperature of the substance, and $D$ is the diffusion coefficient. In this case, In this case, the Reynolds number is set to $Re \in \{1000,\,2000\}$, and the diffusion coefficient is set to $D \in \{0.001,\,0.002\}$.

The shear flow dynamics are governed by the Reynolds number $Re$ and the diffusion coefficient $D$. A high Reynolds number results in inertia-dominated shear instabilities, whereas the diffusion coefficient regulates the relative diffusion rates of the momentum and scalar fields.

The initial velocity field is composed of vertically displaced shear layers, characterized by a hyperbolic tangent profile of width $w$, and horizontally modulated vortices, represented by sine waves with exponentially decaying modes:
\begin{equation}
u_x = \sum_{k=1}^{n_{\text{shear}}} \tanh\left( \frac{5(y - y_k)}{w} \right) + \epsilon_x,
\end{equation}
\begin{equation}
u_y = \sum_{k=1}^{n_{\text{shear}}} \sin(n_{\text{blobs}} \pi x) e^{-25w^2 |y - y_k|^2} + \epsilon_y,
\end{equation}
where  $n_{\text{shear}} = 2$, $n_{\text{blobs}} = 2$, and $w = 0.5$. The scalar field $s$ is initially aligned with the shear layer at $t = 0$. $\epsilon_x$ and $\epsilon_y$ denote random noise components to generate different ICs.

\subsection{Solver Property Analysis}
\label{ap:solver_property_verification}

\begin{figure*}[t!]
\vskip 0in
\begin{center}
\centerline{\includegraphics[width=0.75\textwidth]{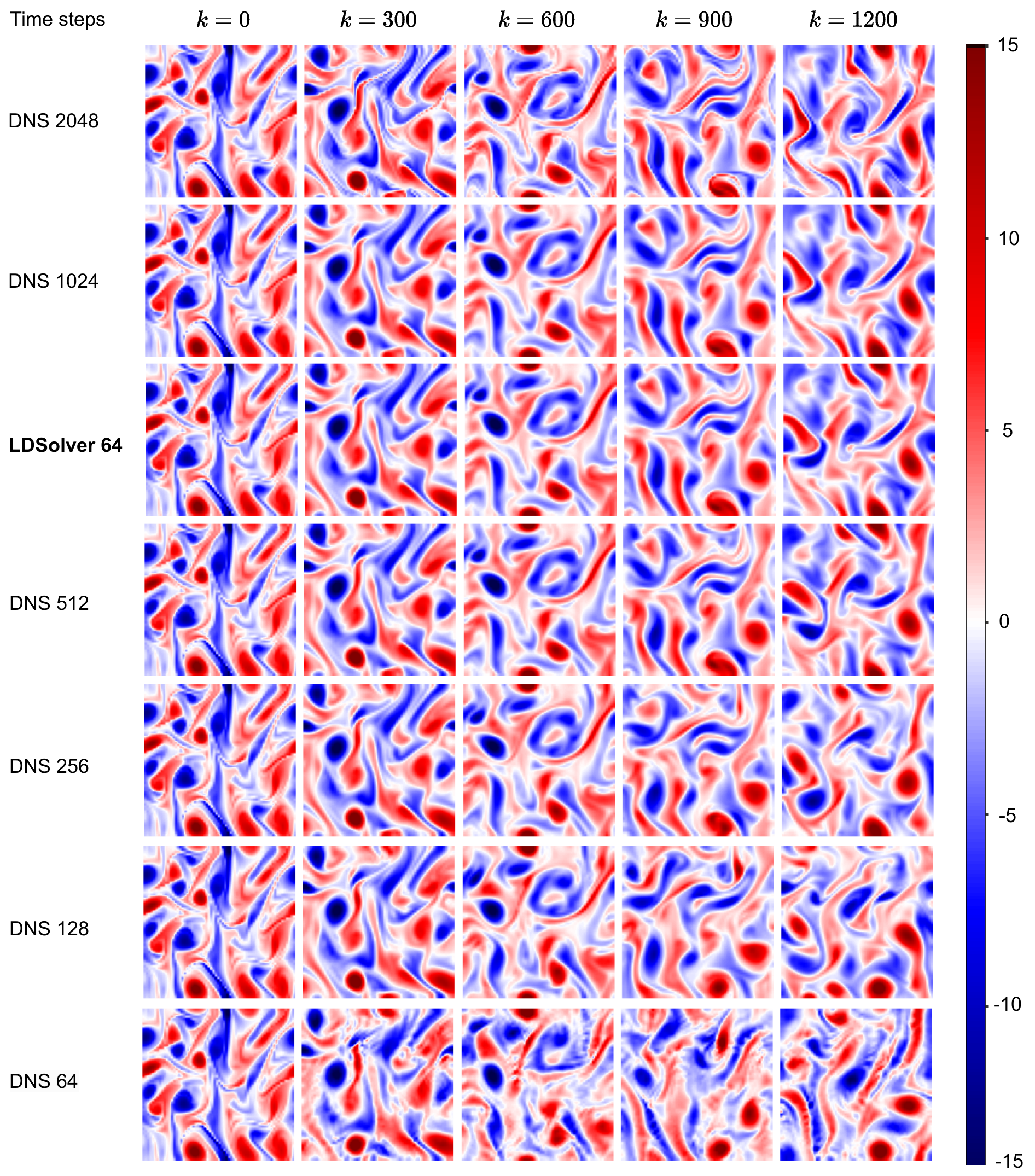}}
\caption{A comparison of vorticity snapshot evolution between DNS solutions at different resolutions and LDSolver at a resolution of 64$\times$64.}
\Description{Side-by-side vorticity snapshots comparing direct numerical simulations (DNS) at various resolutions with LDSolver predictions at 64 by 64 resolution.}
\label{fig:compare DNS snapshot}
\end{center}
\vskip 0in
\end{figure*}

\begin{figure}[t!]
\vskip 0in
\begin{center}
\centerline{\includegraphics[width=0.4\textwidth]{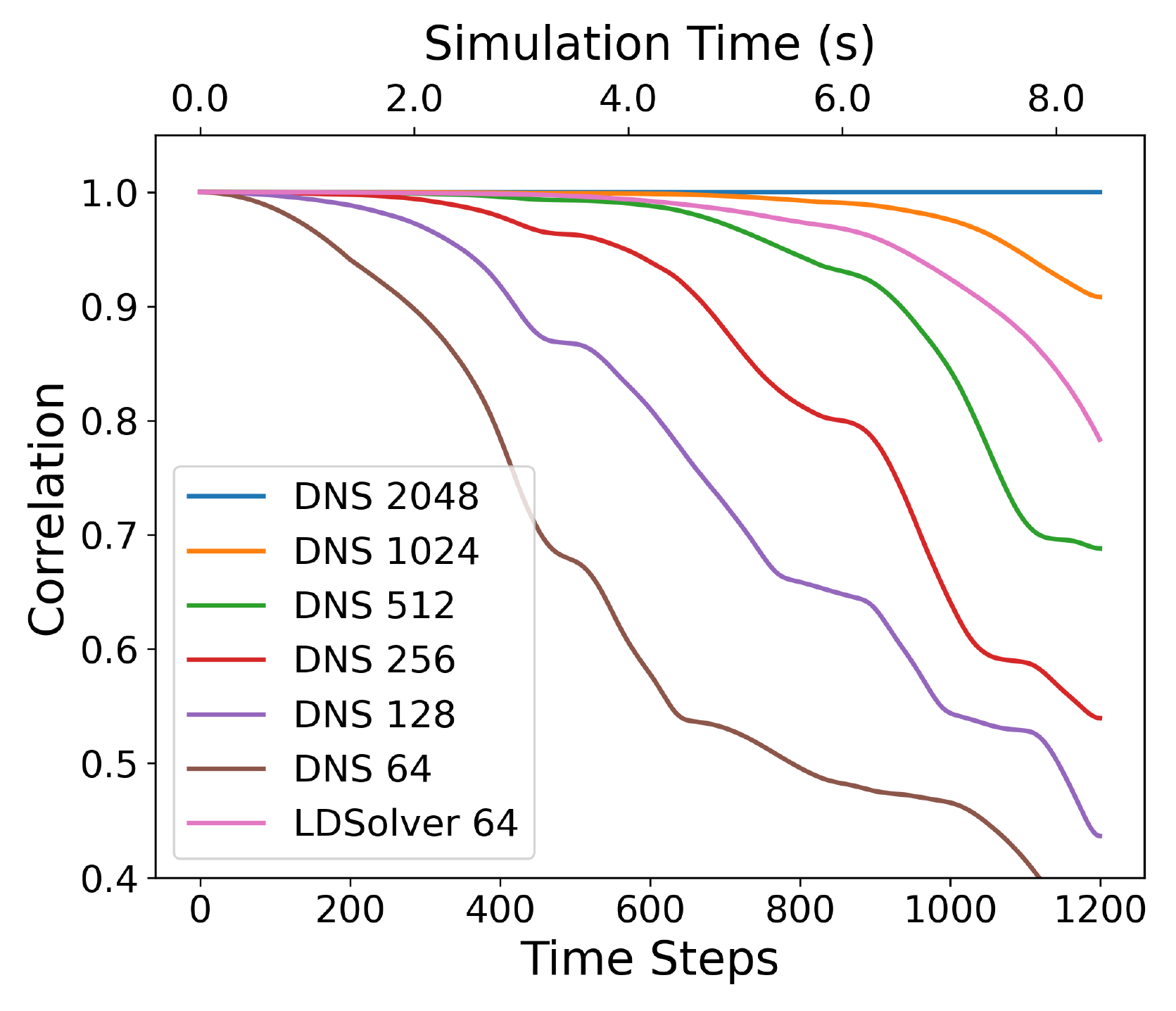}}
\caption{A comparison of vorticity correlation time series between DNS solutions at different resolutions and LDSolver at a resolution of 64$\times$64.}
\Description{Line plot comparing vorticity correlation over time between DNS solutions at multiple resolutions and LDSolver at 64 by 64 resolution.}
\label{fig:compare DNS corrolation}
\end{center}
\vskip 0in
\end{figure}


We analyzed the forced flow at various resolutions. Figure~\ref{fig:compare DNS snapshot} demonstrates that as the resolution decreases, the flow field snapshot gradually degrades. When the resolution is reduced to \(64^2\), the solver's accuracy significantly declines, a trend also reflected in the spatial correlation time series shown in Figure~\ref{fig:compare DNS corrolation}. This systematic analysis underscores the robustness and limitations of our solver across different resolutions. We also observed that LDSolver at a resolution of \(64^2\) achieves higher simulation accuracy than DNS at \(512^2\), approaching the performance of DNS at \(1024^2\).

\section{Training Parameters and Results}

\subsection{Evaluation Metrics}
\label{appendix:metrics}

We adopt multiple metrics to assess the performance of the tested models, including Root Mean Square Error (RMSE), Mean Absolute Error (MAE), Mean Normalized Absolute Difference (MNAD), and High-Correlation Time (HCT)~\cite{Sun2023}. Below, we define these metrics and explain their roles in evaluating prediction accuracy and reliability.
\begin{equation}
\text{RMSE} = \sqrt{\frac{1}{n} \sum_{i=1}^{n} \| \textbf{B}_i - \check{\textbf{B}}_i \|^2}, 
\end{equation}
\begin{equation}
\text{MAE} = \frac{1}{n} \sum_{i=1}^{n} | \textbf{B}_i - \check{\textbf{B}}_i |,
\end{equation}
\begin{equation}
\text{MNAD} = \frac{1}{n} \sum_{i=1}^{n} \frac{\| \textbf{B}_i - \check{\textbf{B}}_i \|}{ \| \textbf{B}_i \|_{\text{max}} - \| \textbf{B}_i \|_{\text{min}}},
\end{equation}
\begin{equation}
\text{HCT} = \sum_{i=1}^{N} \Delta t \cdot\textbf{1} [ \text{PCC}(\textbf{B}_i, \check{\textbf{B}}_i) > 0.8 ],
\end{equation}
\begin{equation}
\text{PCC}(\textbf{B}_i, \check{\textbf{B}}_i) = \frac{\text{cov}(\textbf{B}_i, \check{\textbf{B}}_i)}{\sigma_{\textbf{B}_i} \sigma_{\check{\textbf{B}}_i}},
\end{equation}
where $n$ is the number of trajectories, $\check{\textbf{B}}_i$ denotes the ground truth of the $i$-th trajectory, and $\textbf{B}_i$ represents the predicted spatiotemporal sequence. $\| \cdot \|$ is defined as the Euclidean (or $\ell_2$) norm of a given vector, while $\| \cdot \|_{\text{max}}$ and $\| \cdot \|_{\text{min}}$ denote its maximum and minimum values, respectively. $N$ is the total number of time steps, $\Delta t$ is the time interval, and $\textbf{1}[\cdot]$ is an indicator function returning 1 if the condition holds and 0 otherwise. $\text{cov}(\cdot)$ is the covariance, and $\sigma_{\textbf{B}_i}$, $\sigma_{\check{\textbf{B}}_i}$ are the standard deviations of $\textbf{B}_i$ and $\check{\textbf{B}}_i$, respectively.

Root Mean Square Error (RMSE) quantifies the average magnitude of error between predicted and true values, providing insights into model precision.
Mean Absolute Error (MAE) measures the average absolute deviation between predictions and observations, reflecting the scale of errors. Mean Normalized Absolute Difference (MNAD) evaluates prediction consistency by normalizing errors against the range of observed values. HCT quantifies the model’s capability to maintain reliable long-term predictions. It accumulates the time intervals where the Pearson Correlation Coefficient (PCC) between predictions and ground truth exceeds 0.8.

\begin{table*}[t!]
\centering
\caption{Overview of hyperparameters of LDSolver for Burgers, decaying, forced, and shear flow.}
\vspace{-6pt}
\begin{tabular}{llll}
\toprule
\textbf{Case} & \textbf{Block} & \textbf{Hyperparameters} & \textbf{Values} \\ \hline
\multirow{6}{*}{Burgers}
& \multirow{3}{*}{Physics-based Conv.} & Network & CNN \\ 
& & H$\times$W & 5$\times$4 \\ 
& & Input/Output channel & [1,1] \\ \cline{2-4}
& \multirow{3}{*}{Learnable Conv.} & Network & CNN \\ 
& & H$\times$W & 5$\times$4 \\ 
& & Input/Output channel & [1,1] \\ \hline
\multirow{11}{*}{Decaying Flow}
& \multirow{3}{*}{Physics-based Conv.} & Network & CNN \\ 
& & H$\times$W & 5$\times$4 \\ 
& & Input/Output channel & [1,1] \\ \cline{2-4} 
& \multirow{3}{*}{Learnable Conv.} & Network & CNN \\ 
& & H$\times$W & 5$\times$4 \\ 
& & Input/Output channel & [1,1] \\ \cline{2-4} 
& \multirow{5}{*}{Frequency Domain Operator} & Network & FNO \citep{2021Fourier} \\ 
& & Layers & 4 \\ 
& & Modes & 16 \\ 
& & Width & 8 \\ 
& & $\sigma$ & GELU \\ \hline
\multirow{16}{*}{Forced Flow}
& \multirow{3}{*}{Physics-based Conv.} & Network & CNN \\ 
& & H$\times$W & 5$\times$4 \\ 
& & Input/Output channel & [1,1] \\ \cline{2-4} 
& \multirow{3}{*}{Learnable Conv.} & Network & CNN \\ 
& & H$\times$W & 5$\times$4 \\ 
& & Input/Output channel & [1,1] \\ \cline{2-4} 
& \multirow{5}{*}{Frequency Domain Operator} & Network & FNO \citep{2021Fourier} \\ 
& & Layers & 6 \\ 
& & Modes & 32 \\ 
& & Width & 16 \\ 
& & $\sigma$ & GELU \\ \cline{2-4} 
& \multirow{5}{*}{Temporal Correction Block} & Network & FNO \citep{2021Fourier} \\ 
& & Layers & 4 \\ 
& & Modes & 32 \\ 
& & Width & 8 \\ 
& & $\sigma$ & GELU \\ \hline
\multirow{11}{*}{Shear Flow}
& \multirow{3}{*}{Physics-based Conv.} & Network & CNN \\ 
& & H$\times$W & 5$\times$4 \\ 
& & Input/Output channel & [1,1] \\ \cline{2-4} 
& \multirow{3}{*}{Learnable Conv.} & Network & CNN \\ 
& & H$\times$W & 5$\times$4 \\ 
& & Input/Output channel & [1,1] \\ \cline{2-4} 
& \multirow{5}{*}{\makecell{Frequency Domain Operator}} & Network & FNO \citep{2021Fourier} \\
& & Layers & 4 \\ 
& & Modes & 32 \\ 
& & Width & 8 \\ 
& & $\sigma$ & GELU \\ 
\bottomrule
\end{tabular}
\label{tab:network_hyperparameters}
\end{table*}

\subsection{Baseline models}
\label{appendix:baseline models}
\label{appendix:fno}

\begin{figure*}[t!]
\vskip 0in
\begin{center}
\centerline{\includegraphics[width=0.7\textwidth]{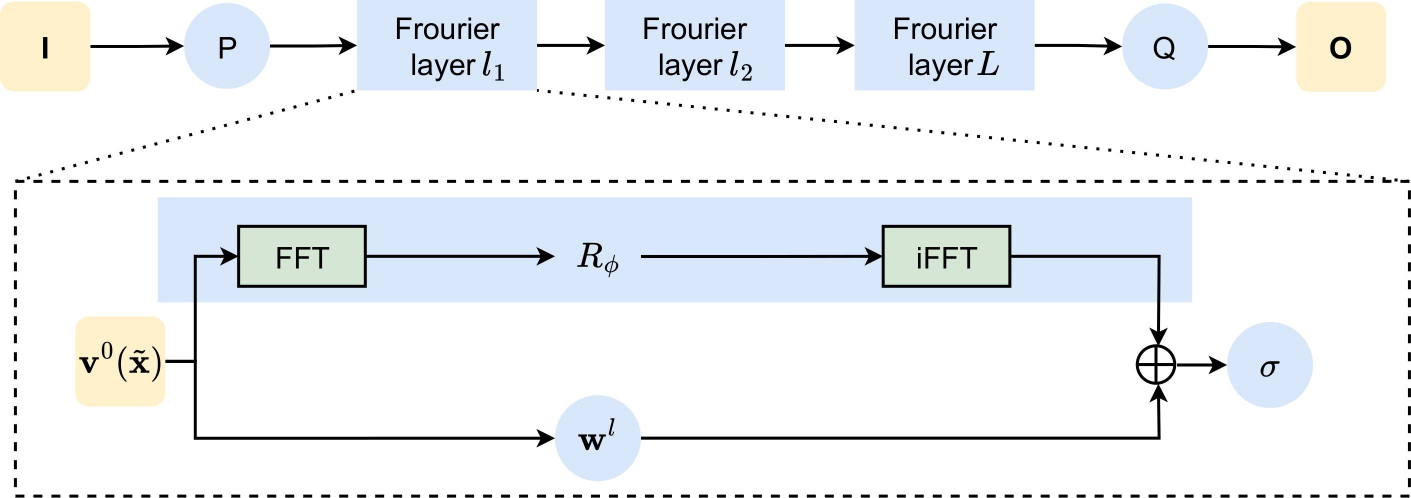}}
\caption{Architecture of the Fourier Neural Operator (FNO) model.}
\Description{Diagram showing the architecture of the Fourier Neural Operator model, including Fourier transforms, lifting and projection layers, and iterative kernel convolution blocks.}
\label{fig:fno model}
\end{center}
\vskip 0in
\end{figure*}

\textbf{Fourier Neural Operator (FNO).} FNO~\cite{2021Fourier} integrates neural networks with Fourier transforms to effectively capture both global and local features of system dynamics.As shown in Figure~\ref{fig:fno model}, the network structure mainly consists of three components: P (lift operation), Q (projection operation), and Fourier layers. Both P and Q are convolutional operations for channel transformation. Each Fourier layer uses the Fast Fourier Transform (FFT) and the inverse Fast Fourier Transform (iFFT) for frequency domain transformations. Additionally, $R_\phi$ represents spectral filtering and convolution in the frequency domain, and $W_l$ denotes the local linear transformation for the $l$-th layer. $\sigma$ is the GeLU activation function.

\newcommand{\cmark}{\ding{51}} 
\newcommand{\xmark}{\ding{55}} 
\begin{table*}[h!]
\caption{Summary analysis of LDSolver and baselines for the NSE example. Each model is assessed based on the listed criteria, with \cmark indicating support and \xmark indicating lack of support.}
\label{tab:baseline_compare}
\vspace{-8pt}
\begin{center}
\begin{small}
\begin{tabular}{lccccccc}
\toprule
Model    & \parbox{1.8cm}{\centering Learnable \& \\ Differentiable}  & \parbox{1.9cm}{\centering Physics \\ Consistency} &\parbox{1.9cm}{\centering Preserves \\ Conservation} & \parbox{1.9cm}{\centering Handles \\ Limited Data} & \parbox{2.3cm}{\centering Rollout \\ Prediction Stability} & \parbox{2cm}{\centering Independent \\ of Historical Data} & \parbox{1.9cm}{\centering Temporal \\ Error Correction} \\
\midrule
FNO      & \cmark   & \xmark& \xmark& \xmark &  \cmark & \cmark &  \xmark \\
LI       & \cmark   & \cmark& \cmark& \xmark  & \xmark  & \cmark  & \xmark  \\
TSM      & \cmark   & \cmark& \cmark & \xmark  & \xmark  & \xmark  & \cmark  \\
PeRCNN   & \cmark   & \cmark& \xmark& \xmark  & \xmark  & \cmark  & \xmark  \\
P$^2$C$^2$Net & \cmark  & \cmark& \xmark& \cmark  & \cmark  & \cmark  & \cmark  \\
\textbf{LDSolver} & \textcolor{blue}{\cmark} & \textcolor{blue}{\cmark} & \textcolor{blue}{\cmark} & \textcolor{blue}{\cmark} & \textcolor{blue}{\cmark} & \textcolor{blue}{\cmark} & \textcolor{blue}{\cmark} \\
\bottomrule
\end{tabular}
\end{small}
\end{center}
\vspace{-6pt}
\end{table*}


\textbf{PeRCNN.} PeRCNN~\cite{Rao2023} embeds physical principles directly into the learning framework by incorporating governing equations within the neural network structure. This architecture features multiple parallel convolutional neural networks (CNNs), which model polynomial relationships through feature map multiplications. The inclusion of physical laws enhances the model's generalization and extrapolation capabilities, enabling accurate predictions in dynamic systems governed by complex equations.

\noindent\textbf{DeepONet.} DeepONet~\cite{Lu2018} is designed to approximate operators and directly map inputs to outputs using neural networks. The architecture consists of two main components: the trunk net, which processes domain-specific information, and the branch net, which handles input functions. This dual-structure approach enables the efficient learning of complex functional relationships and enhances the model's ability to capture detailed operator mappings in various applications.


\noindent\textbf{Temporal Stencil Modeling (TSM).} TSM~\cite{Sun2023} addresses time-dependent partial differential equations (PDEs) in conservation form by integrating time-series modeling with learnable stencil techniques. This method effectively recovers information lost during downsampling, improving predictive accuracy. TSM is particularly beneficial for machine learning models working with coarse-resolution datasets.

\noindent\textbf{P$^2$C$^2$Net.} P$^2$C$^2$Net~\cite{wang2024p} adopts a physics-encoded variable correction learning approach that embeds PDE structures into coarse grids to solve nonlinear dynamic systems. The model introduces a convolutional filter with symmetry constraints, to adaptively compute derivatives of system state variables. It improves the model's ability to capture the characteristics of nonlinear dynamics with limited data, enhancing both generalization and computational efficiency.

\subsection{Descriptive Analysis of Models}
\label{ap:baseline compare}
We analyze the characteristics of LDSolver and existing baselines, focusing on their performance on the NSE example. Table~\ref{tab:baseline_compare} summarizes these methods, highlighting key aspects such as learnability, differentiability, strict physical consistency, handling limited data, stability of rollout predictions, independence from historical data, and temporal error correction. Notably, our proposed model, \text{LDSolver}, excels across all evaluated criteria.

\subsection{Training Details}\label{Training Details}
\label{ap:training details}
All experiments, including training and inference, were conducted on a single Nvidia A100 GPU (80GB memory), running on a server with an Intel(R) Xeon(R) Platinum 8380 CPU (2.30 GHz, 64 cores). Model training was performed on coarse grids (see Table ~\ref{tab:dataset details}).

\textbf{LDSover}: The LDSover architecture employs the Adam optimizer with a learning rate of \( 1 \times 10^{-4} \). The model is trained for 5000 epochs with a batch size of 20. Detailed hyperparameters are listed in Tables~\ref{tab:network_hyperparameters}.

\textbf{P$^2$C$^2$Net}: The P$^2$C$^2$Net architecture \cite{wang2024p} also uses the Adam optimizer with a learning rate of \( 5 \times 10^{-3} \). Training is performed over 5000 epochs, with a batch size of 16. The hyperparameter configurations adhere to the default settings specified in the original paper~\cite{wang2024p}.

\textbf{FNO}: The architecture of the FNO network follows the original study \cite{2021Fourier}, with the main adjustment being the adaptation of the training method to an autoregressive framework. Adam optimizer is used with a learning rate of \( 1 \times 10^{-3} \) and a batch size of 20. Training spans 5000 epochs, with the rollout timestep matching that of LDSover.

\textbf{DeepONet}: The DeepONet model uses its default configuration \cite{Lu2018} and the Adam optimizer. The learning rate is set to \( 5 \times 10^{-4} \), with a decay factor of 0.9 applied every 5000 steps. The model is trained for 5000 epochs with a batch size of 20.

\textbf{PeRCNN}: The PeRCNN model \cite{Rao2023} is used with the standard settings. The optimization is performed using the Adam optimizer with a StepLR scheduler, which reduces the learning rate by a factor of 0.96 every 100 steps. The initial learning rate is set to 0.02, and the model is trained for 1000 epochs with a batch size of 32.


\textbf{TSM}: The TSM architecture \cite{sun2023neural} is used with default settings. The initial learning rate is \( 1 \times 10^{-4} \), and weight decay is set to \( 1 \times 10^{-4} \). The gradient clipping norm is \( 1 \times 10^{-2} \). Adam optimizer is used with \( \beta_2 = 0.98 \), and the batch size is set to 8.

\section{Additional Results}
\label{ap:additional results}

\begin{figure*}[t]
\centering
\vspace{-6pt}
\includegraphics[width=0.95\textwidth]{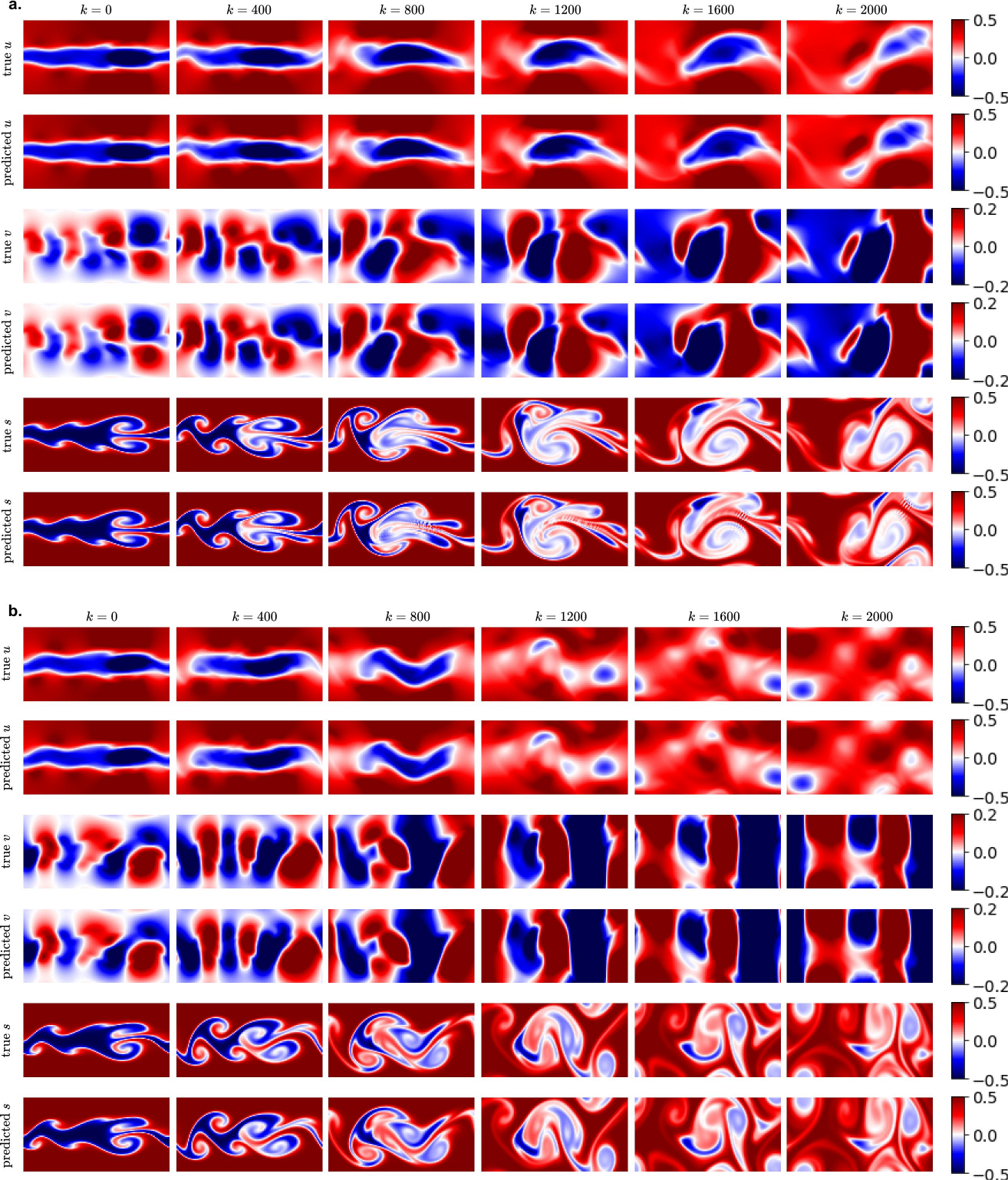} 
\caption{Generalization performance of shear flow. Time steps $k$ are shown horizontally: (a) Trained on $Re=1000$, $D=0.001$ and tested on $Re=2000$, $D=0.001$; (b) Trained on $Re=1000$, $D=0.001$ and tested on $Re=1000$, $D=0.002$. The ground truth is obtained from DNS at a resolution of $2048 \times 1024$, while the predictions are from LDSolver at a resolution of $128 \times 64$.}
\Description{Comparison of LDSolver predictions and DNS ground truth on shear flow generalization tasks. Time-series snapshots for different Reynolds numbers and diffusion coefficients.}
\label{fig:shear_generation}
\end{figure*}

We evaluate LDSolver's generalization capability across both forced and shear flow configurations under varying initial conditions (ICs), Reynolds numbers ($Re$), and diffusion coefficients ($D$). Figure~\ref{fig:shear_generation} demonstrates consistent performance preservation across these parameter variations for the shear flow.

\end{document}